%% file: main.tex
\documentclass{article}

\usepackage[final]{neurips_2021}

\usepackage[utf8]{inputenc} % allow utf-8 input
\usepackage[T1]{fontenc}    % use 8-bit T1 fonts
\usepackage{hyperref}       % hyperlinks
\usepackage{url}            % simple URL typesetting
\usepackage{booktabs}       % professional-quality tables
\usepackage{amsfonts}       % blackboard math symbols
\usepackage{nicefrac}       % compact symbols for 1/2, etc.
\usepackage{microtype}      % microtypography

\usepackage[table]{xcolor}
\usepackage{graphicx}
\usepackage{makecell}
\usepackage{multirow}
\usepackage{bm}

\usepackage{amsmath}
\usepackage{amssymb}
\usepackage{chessfss}
\usepackage{algorithm}
\usepackage[noend]{algorithmic}

\usepackage[thinlines]{easytable}

\usepackage{tikz}
\usetikzlibrary{positioning,decorations.markings,calc}

\input{setup.tex}

\input{chessboard.tex}
\title{Deep Synoptic Monte Carlo Planning in Reconnaissance Blind Chess}

\author{%
  Gregory Clark \\
  ML Collective, Google \\
  \texttt{gregoryclark@google.com} \\
}

\begin{document}

\maketitle

\begin{abstract}
This paper introduces deep synoptic Monte Carlo planning (DSMCP)
for large imperfect information games.
The algorithm constructs a belief state with an unweighted particle filter
and plans via \playouts{} that start at samples
drawn from the belief state.
The algorithm accounts for uncertainty by performing inference on ``synopses,''
a novel stochastic abstraction of \longinfostates{}.
DSMCP is the basis of the program \penumbra{}, which won
the official 2020 reconnaissance blind chess competition
versus 33 other programs.
This paper also evaluates algorithm variants
that incorporate caution, paranoia, and a novel bandit algorithm.
Furthermore, it audits the synopsis features used in \penumbra{}
with per-bit saliency statistics.
\end{abstract}

\section{Introduction}
\label{sec:introduction}

Choosing a Nash equilibrium strategy
is rational when the opponent is able to
identify and exploit suboptimal behavior \citep{bowling2001rational}.
However, not all opponents are so responsive,
and computing a Nash equilibrium is intractable for many games.
This paper presents deep synoptic Monte Carlo planning (DSMCP), an
algorithm for large imperfect information games that
seeks a best-response strategy rather than a Nash equilibrium strategy.

When opponents use fixed policies,
an imperfect information game may be viewed as a
partially observable Markov decision process (POMDP)
with the opponents as part of the environment.
DSMCP treats playing against specific opponents as related offline
reinforcement learning (RL) problems and exploits predictability.
Importantly,
the structure of having opponents with imperfect information is preserved
in order to account for their uncertainty.

\begin{figure}[t]
\begin{minipage}{.49\linewidth}
\input{first_figure.tex}
\end{minipage}%
\begin{minipage}{0.02\linewidth}
\,
\end{minipage}%
\begin{minipage}{.49\linewidth}
\begin{center}
\begin{minipage}{0.5\columnwidth}
    \centering
    \includegraphics[width=0.96\linewidth]{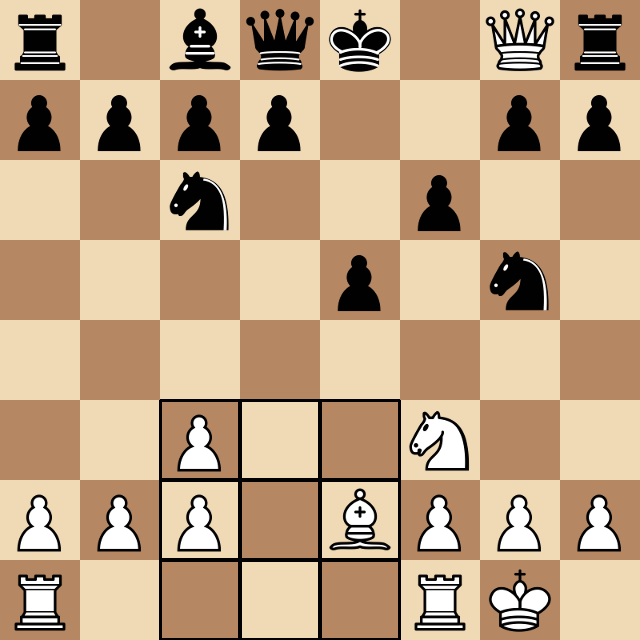}
    (a)
\end{minipage}%
\begin{minipage}{0.5\columnwidth}
    \centering
    \includegraphics[width=0.96\linewidth]{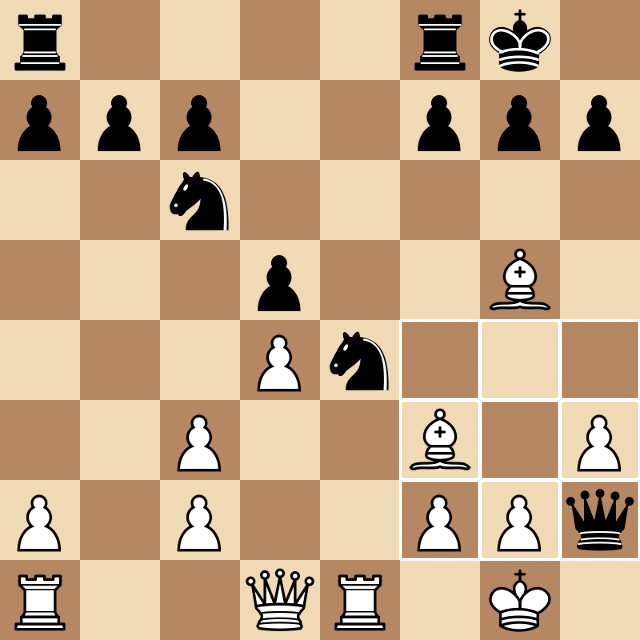}
    (b)
\end{minipage}%
\end{center}
\caption{
Playing RBC well requires balancing risks and rewards.
(a) On the left, \penumbra{} moved the white queen to \texttt{g8}.
After sensing at \texttt{d2}, Black could infer that the white queen
occupied one of 25 squares.
That uncertainty allowed the white queen
to survive and capture the black king on the next turn.
(b) On the right, \penumbra{} moved the black queen to \texttt{h2}.
In this case, the opponent detected and captured the black queen.
The games are available online at
\url{https://rbc.jhuapl.edu/games/120174} and
\url{https://rbc.jhuapl.edu/games/124718}.
}
\label{fig:queen-risks}
\end{minipage}
\end{figure}

DSMCP uses sampling to break the
``curse of dimensionality'' \citep{pineau2006anytime} in three ways:
sampling possible histories with a particle filter,
sampling possible futures with
upper confidence bound tree search (UCT) \citep{kocsis2006bandit},
and sampling possible \latentstates{} within each \longinfostate{} uniformly.
It represents \longinfostate{}s with a generally-applicable
stochastic abstraction technique that
produces a ``synopsis'' from sampled \latentstates{}.
This paper assesses DSMCP on reconnaissance blind chess (RBC),
a large imperfect information chess variant.

\section{Background}
\label{sec:related-work}

Significant progress has been made in recent years in
both perfect and imperfect information settings.
For example, using deep neural networks to guide UCT
has enabled monumental achievements in
abstract strategy games as well as computer games
\citep{silver2016mastering, silver2017mastering,
       silver2017chess, schrittwieser2019mastering,
       wu2020accelerating, tomasev2020assessing}.
This work employs deep learning in a similar fashion. 

Recent advancements in imperfect information games are also remarkable.
Several programs have reached superhuman performance in Poker
\citep{moravcik2017deepstack, brown2018superhuman, brown2019superhuman, brown2020combining}.
In particular, ReBeL \citep{brown2020combining}
combines RL and search by converting imperfect information games
into continuous state space perfect information games
with public belief states as nodes.
This approach is powerful, but it relies on public knowledge and fails to scale to
games with hidden actions and substantial private information, such as RBC.

Information set search \citep{parker2006overconfidence, parker2010paranoia}
is a limited-depth algorithm for imperfect information games
that operates on \longinfostates{} according to a minimax rule.
This algorithm was designed for and evaluated
on Kriegspiel chess, which is comparable to RBC.

Partially observable Monte Carlo planning (POMCP) \citep{silver2010pomcp}
achieves optimal policies for POMDPs
by tracking approximate belief states with an unweighted particle filter
and planning with a variant of UCT on a search tree of histories.
In practice, POMCP can suffer from particle depletion,
requiring a domain-specific workaround.
This work combines an unweighted particle filter
with a novel \longinfostate{} abstraction technique
which increases sample quality and supports deep learning.

Smooth UCT \citep{heinrich2015smoothuct} and
information set Monte Carlo tree search (ISMCTS) \citep{cowling2012ismcts}
may be viewed as multi-agent versions of POMCP.
These two algorithms for playing extensive-form games
build search trees (for each player) of \longinfostates{}.
These two algorithms and DSMCP all
perform playouts from determinized states
that are accurate from the current player's perspective,
effectively granting the opponent extra information.
Still, Smooth UCT approached a Nash equilibrium
by incorporating a stochastic bandit algorithm into its tree search.
DSMCP uses a similar bandit algorithm that mixes in
a learned policy during early node visits.

While adapting perfect information algorithms has performed surprisingly well
in some imperfect information settings \citep{long2010pimc},
the theoretical guarantees of
variants of counterfactual regret minimization (CFR)
\citep{neller2013introduction, brown2018deep}
are enticing.
Online outcome sampling (OOS) \citep{lisy2015online}
extends Monte Carlo counterfactual regret minimization (MCCFR) \citep{lanctot2009regret}
by building its search tree incrementally and
targeting playouts to relevant parts of the tree.
OOS draws samples from the beginning of the game.
MCCFR and OOS are theoretically guaranteed to converge
to a Nash equilibrium strategy.
Specifically, CFR-based algorithms produce mixed strategies
while DSMCP relies on incidental stochasticity.

Neural fictitious self-play (NFSP) \citep{heinrich2016deep} is an RL algorithm
for training two neural networks for imperfect information games.
Experiments with NFSP employed compact observations embeddings of \longinfostates{}.
DSMCP offers a generic technique for embedding \longinfostates{} in large games.
Dual sequential Monte Carlo (DualSMC) \citep{wang2019dualsmc}
estimates belief states and plans
in a continuous domain via sequential Monte Carlo with heuristics.

\section{Reconnaissance blind chess}
\label{sec:reconnaissance-blind-chess}

Reconnaissance blind chess (RBC)
\citep{newman2016reconnaissance, markowitz2018complexity, pmlr-v123-gardner20a}
is a chess variant that incorporates
uncertainty about the placement of the opposing pieces
along with a private sensing mechanism.
As shown in Figure~\ref{fig:log-information-set-graph},
RBC players are often faced with thousands of possible game states,
and reducing uncertainty increases the odds of winning.

\paragraph{Game rules}

Pieces move in the same way in RBC as in chess.
Players cannot directly observe the movement of the opposing pieces.
However, at the beginning of each turn,
players may view the ground truth of any $3\stimes3$ patch of the board.
The information gained from the sensing action remains private to that player.
Players are also informed of the location of all captures,
but not the identity of capturing pieces.
When a requested move is illegal,
the move is substituted with the closest legal move
and the player is notified of the substitution.
For example, in Figure~\ref{fig:queen-risks}~(a),
if Black attempted to move the rook from \texttt{h8} to \texttt{f8},
the rook would capture the queen on \texttt{g8} and stop there instead.
Players are always able to track the placement of their own pieces.
Capturing the opposing king wins the game, and
players are not notified about check.
Passing and moving into check are legal.

\paragraph{Official competition}

This paper introduces the program \penumbra{},
the winner of the official 2020 RBC rating competition.
In total, 34 programs competed to achieve the highest rating
by playing public games.
Ratings were computed with \textit{BayesElo} \citep{coulom2008whr},
and playing at least 100 games was required to be eligible to win.
Figure~\ref{fig:queen-risks} shows ground truth positions from the tournament in which \penumbra{} voluntarily put its queen in danger.
Players were paired randomly,
but the opponent's identity was provided at the start of each game
which allowed catering strategies for specific opponents.
However, opponents were free to change their strategies at any point,
so attempting to exploit others could backfire.
Nonetheless, \penumbra{} sought to model and counter predictable opponents
rather than focusing on finding a Nash equilibrium.

\paragraph{Other RBC programs}

RBC programs have employed a variety of
algorithms \citep{pmlr-v123-gardner20a} including
Q-learning \citep{mnih2013playing},
counterfactual regret minimization (CFR) \citep{zinkevich2008regret},
online outcome sampling (OOS) \citep{lisy2015online},
and the Stockfish chess engine \citep{romstad2020stockfish}.
Another strong RBC program
\citep{highley2020, blowitski2021checkpoint}
maintains a probability distribution for each piece.
Most RBC programs select sense actions and move actions in separate ways
while DSMCP unifies all action selection.
\citet{savelyev2020mastering} also applied UCT to RBC and modeled the
root belief state with a distribution over 10,000 tracked positions.
Input to a neural network consisted of the most-likely 100 positions,
and storing a single training example required
3.5MB on average,
large enough to hinder training.
This work circumvented the same issue by
representing training examples with compact synopses which are
less than 1kB.
% without compression.

\section{Terminology}
\label{sec:extensive-form-games}

Consider the two-player extensive-form game with
agents $\mathcal{P} = \{$self, opponent$\}$, actions $\mathcal{A}$,
``ground truth'' \latentstates{} $\latstates{}$,
and initial state $\latstate_0 \in \latstates{}$.
Each time an action is taken,
each agent $p \in \mathcal{P}$ is given an observation
$\observe_p \in \mathcal{O}$ that matches ($\sim$) the possible \latentstates{} from $p$'s perspective.
For simplicity, assume the game has deterministic actions
such that each $a \in \mathcal{A}$ is a function $a : \latstateset{} \rightarrow \latstates{}$
defined on a subset of \latentstates{} $\latstateset{} \subset \latstates{}$.
Define $\mathcal{A}_\latstate{}$ as the set of actions available from $\latstate{} \in \latstates{}$.

An \longinfostate{} (\infostate{}) $s \in \mathcal{S}$ for agent $p$
consists of all observations $p$ has received so far.\footnote{
An \infostate{} is equivalent to an information set,
which is the set of all possible action histories from $p$'s perspective
\citep{osborne1994course}.
}
Let $\latstates{}_s \subset \latstates{}$ be the set of all
\latentstates{} that are possible from $p$'s perspective from $s$.
In general, $\latstates_s$ contains less information than $s$
since some (sensing) actions may not affect the world state.
Define a collection of limited-size \latentstate{} sets
$\mathcal{L} = \{L \subset \latstates_s : s \in \mathcal{S}, |L| \le \ell\}$,
given a constant $\ell$.

Let
$\rho : \latstates{} \rightarrow \mathcal{P}$ indicate the agent to act in each \latentstate{}.
Assume
that $\mathcal{A}_\latstate{} = \mathcal{A}_y$ and $\rho(\latstate{}) = \rho(y)$
for all $\latstate{}, y \in \latstates{}_s$ and $s \in \mathcal{S}$.
Then extend the definitions of
actions available $\mathcal{A}_*$ and agent to act $\rho$
over sets of \latentstates{} and over \infostates{} in the natural way.
A policy $\pi(a | s)$
is a distribution over actions given an \infostate{}.
A belief state $B(h)$ is a distribution over action histories.
Creating a belief state from an \infostate{}
requires assumptions about the opponent's action policy $\tau(a|s)$.
Let $\mathcal{R}_p : \latstates{} \rightarrow \mathbb{R}$ map terminal states to the reward for player $p$.
Then $(\mathcal{S}, \mathcal{A}, \mathcal{R}_\text{self}, \tau, s_0)$
is a POMDP, where
the opponent's policy $\tau$ provides environment state transitions
and $s_0$ is the initial \infostate{}.
In the rest of this paper, the word ``state'' refers to a \latentstate{}
unless otherwise specified.

\section{Deep synoptic Monte Carlo planning}
\label{sec:dsmcp-algorithm}

\begin{figure}
\begin{minipage}{.5\linewidth}
\begin{center}
\input{overview_figure.tex}
\end{center}
\caption{%
DSMCP approximates \infostates{}
with size-limited sets of possible states (circles).
It tracks all possible states $\stateset_t$ for each turn from its own perspective
and constructs belief states $\beliefstate_t$ with
approximate \infostates{} from the opponent's perspective.
At the root of each \playout{},
the initial approximate \infostate{} for the opponent
is sampled from $\beliefstate_t$, and the initial approximate \infostate{} for
itself is a random subset of $\stateset_t$.
}
\label{fig:algorithm-figure}
\end{minipage}%
\begin{minipage}{.5\linewidth}
\input{algorithms_one.tex}
\end{minipage}
\end{figure}

Effective planning algorithms for imperfect information games
must model agents' choice of actions based on (belief states derived from)
\infostates{}, not on \latentstates{} themselves.
Deep synoptic Monte Carlo planning (DSMCP) approximates \infostates{}
with size-limited sets of possible \latentstates{} in $\mathcal{L}$.
It uses those approximations to construct a belief state and as UCT nodes
\citep{kocsis2006bandit}.
Figure~\ref{fig:algorithm-figure} provides a high-level
visualization of the algorithm.

\begin{figure}
\input{algorithms.tex}
\end{figure}

A bandit algorithm chooses an action during each node visit,
as described in Algorithm~\ref{alg:stochastic-bandit}.
This bandit algorithm is similar to Smooth UCB \citep{heinrich2015smoothuct}
in that they both introduce stochasticity by mixing in a secondary policy.
Smooth UCB empirically approached a Nash equilibrium
utilizing the average policy according to action visits at each node.
DSMCP mixes in a neural network's policy ($\pi$) instead.
The constant $c$ controls the level of exploration,
and $m$ controls how the policy $\pi$ is mixed into the bandit algorithm.
For example, taking $m = 0$ always selects actions directly with $\pi$
without considering visit counts, and taking $m = \infty$ never mixes in $\pi$.
As in \cite{silver2016mastering},
$\pi$ provides per-action exploration values which guide the tree search.

Approximate belief states are constructed as subsets
$\beliefstate \subset \mathcal{L}$, where each $L \in \beliefstate$
is a set of possible world-states from the opponent's perspective.
This ``second order'' representation of
belief states allows DSMCP to account for the opponent's uncertainty.
\Infostates{} sampled with rejection
(Algorithm~\ref{alg:prepare-sample})
are used as the ``particles'' in a particle filter which models
successive belief states.
Sampling is guided by a neural network policy ($\hat{\tau}$)
based on the identity of the opponent.
To counter particle deprivation, if $k$ consecutive candidate samples
are rejected as incompatible with the possible world states,
then a singleton sample consisting of a randomly-chosen possible
state is selected instead.

The tree search, described in Algorithm~\ref{alg:choose-action},
tracks an approximate \infostate{} for each player while simulating \playouts{}.
\Playouts{} are also guided by
policy ($\pi$ and $\hat{\tau}$) and value ($\nu$) estimations from a neural network.
A synopsis function $\sigma$
creates a fixed-size summary of each node as input for the network.
The constant $b$ is the batch size for inference, $d$ is the search depth,
$\ell$ is the size of approximate \infostate{}s,
$n_{\text{vl}}$ is the virtual loss weight,
and $z$ is a threshold for increasing search depth.

Algorithm~\ref{alg:play-game} describes how to play an entire game,
tracking all possible \latentstates{}.
Approximate belief states ($\beliefstate_t$) are constructed for each past turn
by tracking $n_{\text{particles}}$ elements of $\mathcal{L}$
(from the opponent's point of view) with an unweighted particle filter.
Each time the agent receives a new observation,
all of the (past) particles that are inconsistent with the observation
are filtered out and replenished, starting with the oldest belief states.

\subsection{Synopsis}

\input{example_game.tex}

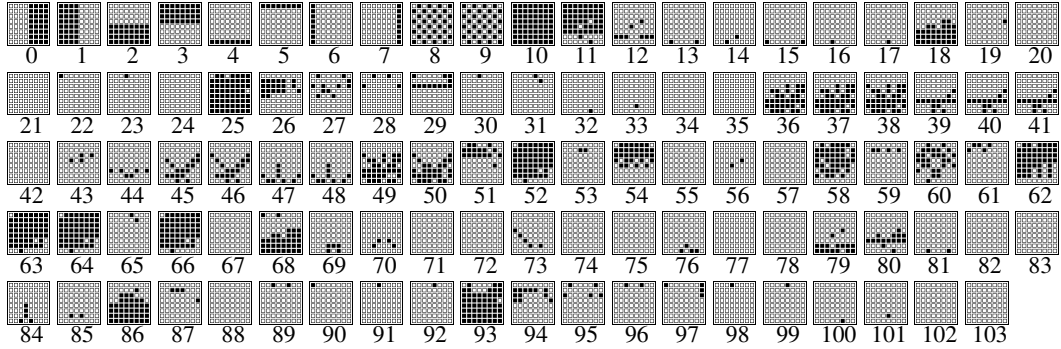
\begin{figure*}[ht]
\begin{center}
\input{bitboards_wide.tex}
\end{center}
\caption{
This set of synopsis bitboards was used as input to the neural network
before White's sense on turn 5 of the game in Figure~\ref{fig:sample-game}.
The synopsis contains 104 bitboards.
Each bitboard encodes 64 binary features
of the possible state set that the synopsis summarizes.
For example, bitboard \#26 contains the possible locations of opposing pawns, and
bitboard \#27 contains the possible locations of opposing knights.
An attentive reader may notice that the black pawn on \texttt{h4} is missing from bitboard \#26,
which is due to subsampling to $\ell = 128$ states before computing the bitboards.
In this case, the true state was missing from the set of states used to create the synopsis.
The features in each synopsis are only approximations of the
\infostates{} that they represent.
The first 10 bitboards are constants, which provide information that
is difficult for convolutions to construct otherwise
\citep{liu2018coordconv}.
}
\label{fig:sample-situation}
\end{figure*}

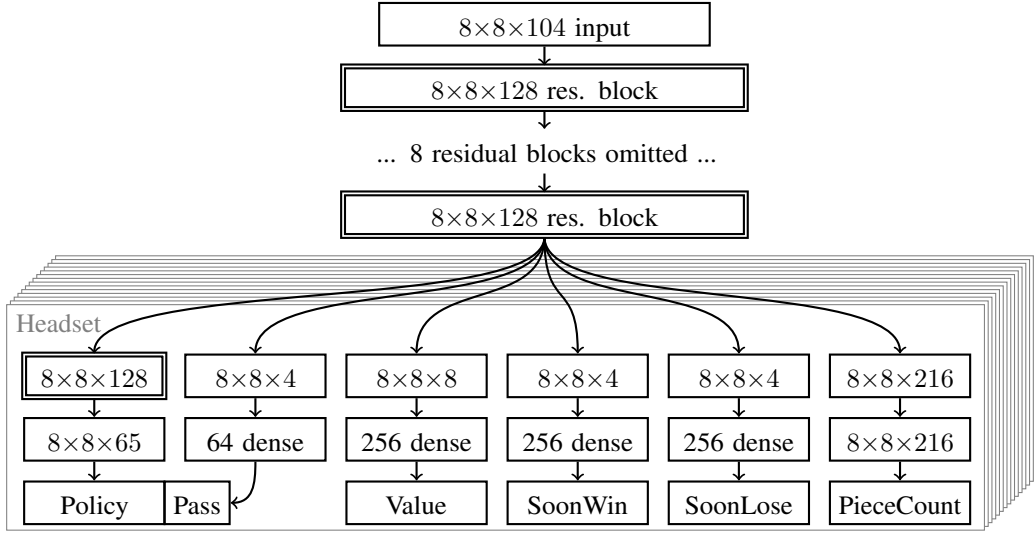
\begin{figure*}[ht]
\begin{center}
\input{architecture.tex}
\end{center}
\caption{
\penumbra{}'s network contains a shared tower with 10 residual blocks
and 14 headsets.
Each headset contains 5 heads for a total of 70 output heads.
The residual blocks are shown with a double border,
and they each contain two $3 \stimes 3$ convolutional layers and batch normalization.
All of the convolutional layers in the headsets are $1 \stimes 1$ convolutions
with the exception of the one residual block for each policy head.
Each headset was trained on a separate subset of the data, as described in
Table~\ref{tab:headset-descriptions}.
The policy head provides logits for both sense and move actions.
}
\label{fig:architecture-diagram}
\end{figure*}

One of the contributions of this paper is the methodology
used to approximate and encode \infostates{}.
Games that consist of a fixed number of turns, such as poker, admit a
naturally-compact \infostate{} representation based on observations
\citep{heinrich2015smoothuct, heinrich2016deep}.
However, perfect representations are not always practical.
Game abstractions are often used to reduce
computation and memory requirements.
For example, imperfect recall is an effective abstraction when past actions are
unnecessary for understanding the present situation
\citep{waugh2009imperfectrecall, lanctot2012noregret}.

DSMCP employs a stochastic abstraction
which represents \infostates{} with sets of \latentstates{}
and then subsamples to a manageable cardinality $(\ell)$.
Finally, a permutation-invariant synopsis function $\sigma$ produces
fixed-size summaries of the approximate \infostates{}
which are used for inference.
An alternative is to run inference on ``determinized'' \latentstates{} individually
and then somehow aggregate the results.
However, such aggregation can easily lead to strategy fusion \citep{frank1998finding}.
Other alternatives include evaluating states with a recurrent network
\citep{rumelhart1986learning}
one-at-a-time or using a permutation-invariant architecture
\citep{zaheer2017deep, wagstaff2019limitations}.

Given functions $\featfunc_i : \latstates \rightarrow \{0, 1\}$
for $i = 0, \dots, \numfeatures$
that map states to binary features,
define the $i^\text{th}$ component of a synopsis function
$\sigma : \mathcal{L} \rightarrow \{0, 1\}^{\numfeatures}$
as
\begin{equation}
    \sigma_i(\latstateset) =
    \featfunc_i(\latstate_0) *_i \featfunc_i(\latstate_1) \ast_i \dots *_i \featfunc_i(\latstate_{\ell})
\end{equation}
where $\latstateset = \{\latstate_0, \latstate_1, \dots, \latstate_\ell\}$
and $*_i$ is either the logical \texttt{AND} ($\land$)
or the logical \texttt{OR} ($\lor$) operation.
For example, if $\featfunc_i$ encodes whether 
an opposing knight can move to the \texttt{d7} square of a chess board
and $*_i = \land$, then
$\sigma_i$ indicates that a knight can definitely move to \texttt{d7}.
Figure~\ref{fig:sample-game} shows an example game, and
Figure~\ref{fig:sample-situation} shows an example
output of \penumbra{}'s synopsis function,
which consists of 104 bitboard feature planes each with 64 binary features.
The appendix describes each feature plane.

\subsection{Network architecture}

\penumbra{} uses a residual neural network
\citep{he2016resnet}
as shown in Figure~\ref{fig:architecture-diagram}.
The network contains 14 headsets,
designed to model specific opponents and regularize each other
as they are trained on different slices of data \citep{zhang2020balance}.
Each headset contains 5 heads:
a policy head, a value head, two heads for predicting
winning and losing within the next 5 actions,
and a head for guessing the number of pieces of each type in the
ground truth \latentstate{}.
The \focustop{} policy head % ($\pi$)
and the \focusall{} value head % ($\nu$)
are used for planning as $\pi$ and $\nu$, respectively.
The other heads
(including the \texttt{SoonWin}, \texttt{SoonLose}, and \texttt{PieceCount} heads)
provide auxiliary tasks for further
regularization \citep{wu2020accelerating, fifty2020measuring}.
While playing against an opponent that is ``recognized''
(when a headset was trained on data from only that opponent),
the policy head ($\hat{\tau}$) of the corresponding headset is used
for the opponent's moves while progressing the particle filter
(Algorithm~\ref{alg:prepare-sample}) and
while constructing the UCT tree (Algorithm~\ref{alg:choose-action}).
When the opponent is unrecognized, the \focustop{} policy head is used by default.

\subsection{Training procedure}
\label{sec:training-procedure}

The network was trained on historical%
\footnote{%
The games were downloaded
from \url{rbmc.jhuapl.edu} in June, 2019 and \url{rbc.jhuapl.edu} in August, 2020.
Additionally, 5,000 games were played locally by \focusstockyinference{}.
} game data
as described by Table~\ref{tab:headset-descriptions}.
The reported accuracies are averages over 5 training runs.
The \focusall{} headset was trained on all games,
the \focustop{} headset was trained on games from the highest-rated players,
the \focushuman{} headset was trained on all games played by humans,
and each of the other 11 headsets were trained to mimic specific opponents.

10\% of the games were used as validation data based on game filename hashes.
Training examples were extracted from games multiple times since
reducing possible state sets to $\ell$ states is non-deterministic.
A single step of vanilla stochastic gradient descent
was applied to one headset at a time,
alternating between headsets according to their training weights.
See the appendix for hyperparameter settings and accuracy cross tables.
Training and evaluation were run on four RTX 2080 Ti GPUs.

\input{headset_summary.tex}

\subsection{Implementation details}
\label{sec:implementation-details}

\penumbra{} plays RBC with DSMCP along with
RBC-specific extensions.
First, sense actions that are dominated by other sense actions are pruned from consideration.
Second, \penumbra{} can detect some forced wins in
the sense phase, the move phase, and during the opponent's turn.
This static analysis is applied at the root and to \playouts{};
\playouts{} are terminated as soon as a player could win,
avoiding unnecessary neural network inference.
The static analysis was also used to clean training games in which
the losing player had sufficient information to find a forced win.

Piece placements are represented with bitboards \citep{browne2014bitboard},
and the tree of approximate \infostates{}
is implemented with a hash table.
Zobrist hashing \citep{zobrist1990hashing}
maintains  hashes of piece placements incrementally.
Hash table collisions are resolved by overwriting older entries.
The tree was not implemented until after the competition, so
fixed-depth \playouts{} were used instead ($m = 0$).
Inference is done in batches of $256$
during both training and online planning.
The time used per action is approximately proportional to the time remaining.
The program processes approximately 4,000 nodes per second, and it
plays randomly when the number of possible states exceeds 9 million.

\section{Experiments}
\label{sec:experiments}

This section presents the results of playing games between \penumbra{} and
several publicly available RBC baselines
\citep{pmlr-v123-gardner20a, bernardoni2020baselines}.
Each variant of \penumbra{} in Table~\ref{tab:baseline-bots} played 1000 games against each baseline, and each variant in 
Table~\ref{tab:caution-and-paranoia}
and Table~\ref{tab:exploration} played 250 games against each baseline.
Games with errors were ignored and replayed.
The Elo ratings and 95\% confidence intervals
were computed with \textit{BayesElo} \citep{coulom2008whr}
and are all compatible.
The scale was anchored with \focusstockyinference{} at 1500
based on its rating during the competition.

Table~\ref{tab:baseline-bots} gives ratings of the baselines and
five versions of \penumbra{}.
\texttt{PenumbraCache}
relied solely on the network policy for action selection in playouts
($m \,{=}\, 0$),
\texttt{PenumbraTree} built a UCT search tree
($m \,{=}\, \infty$), and
\texttt{PenumbraMixture}
mixed in the network policy during early node visits
($m \,{=}\, 1$).
The mixed strategy performed the best.
\texttt{PenumbraNetwork} selected actions
based on the network policy without performing any playouts.
\texttt{PenumbraSimple}
is the same as \texttt{PenumbraMixture}
with the static analysis described
in Section~\ref{sec:implementation-details} disabled.
\texttt{PenumbraNetwork} and \texttt{PenumbraSimple} serve as ablation studies;
removing the search algorithm is detrimental while the effect of removing
the static analysis is not statistically significant.
Unexpectedly, \penumbra{} played the strongest against \focusstockyinference{}
when that program was unrecognized.
So, in this case, modeling the opponent with a stronger policy
outperformed modeling it more accurately.

Two algorithmic modifications that give the opponent
an artificial advantage during planning were investigated.
Table~\ref{tab:caution-and-paranoia}
reports the results of a grid search over
``cautious'' and ``paranoid'' variants of DSMCP.
The caution parameter $\kappa$ specifies the percentage
of \playouts{} that use $\ell = 4$
for the opponent instead of the higher default limit.
Since each approximate \infostate{}
is guaranteed to contain the correct ground truth in \playouts{},
reducing $\ell$ for the opponent gives the opponent higher-quality information,
allowing the opponent to counter risky play more easily in the constructed UCT tree.

The paranoia parameter augments the exploration values in
Algorithm~\ref{alg:stochastic-bandit} to incorporate
the minimum value seen during the current \playout{}.
With paranoia $\phi$, actions are selected according to
\begin{equation}
  \argmax_a \left((1 - \phi)\frac{ \actionvalues_a }{ \actioncounts_a }
             + \phi \actionminimums_a
             + c \pi_a \sqrt{\frac{\ln{\actiontotalcount}}{\actioncounts_a}} \right)
\end{equation}
where $\actionminimums$ contains the minimum value observed for each action.
This is akin to the notion of paranoia studied by
\citet{parker2006overconfidence, parker2010paranoia}.

\input{result_tables.tex}

\begin{figure}[h]
  \centering
  \begin{minipage}{0.5\linewidth}
    \centering
    \includegraphics[width=1\linewidth]{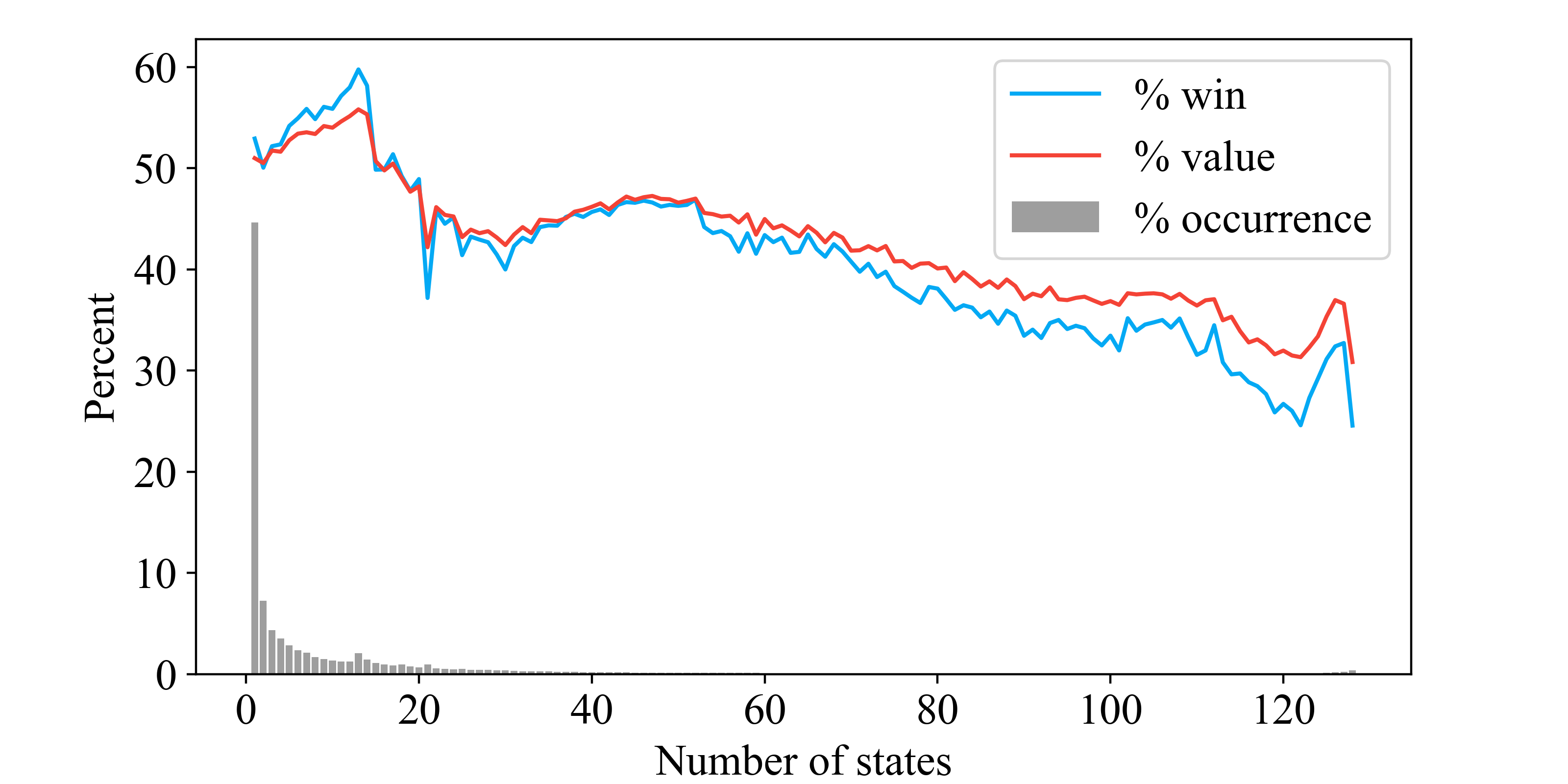}
    (a)
  \end{minipage}%
  \begin{minipage}{0.5\linewidth}
    \centering
    \includegraphics[width=1\linewidth]{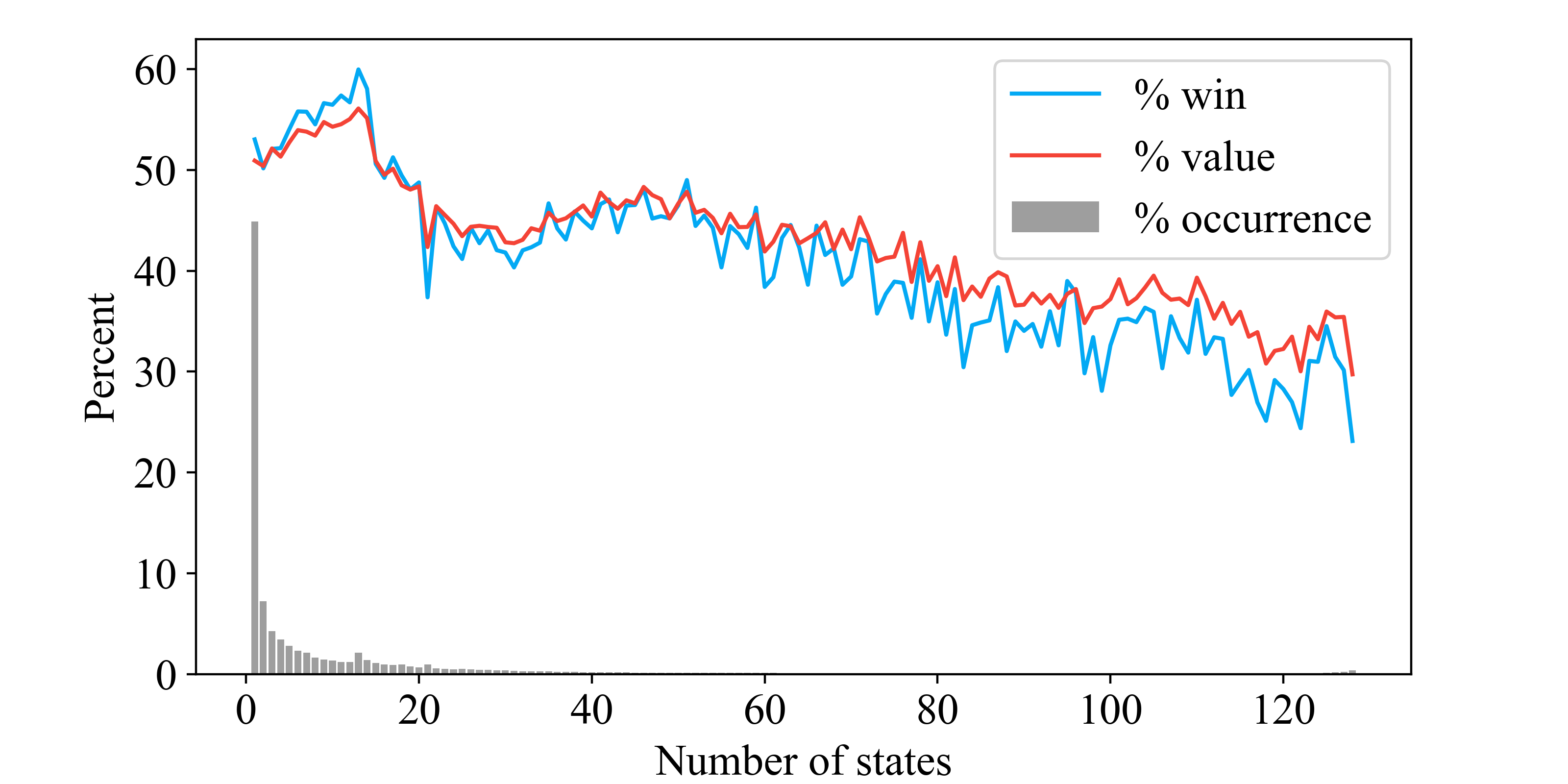}
    (b)
  \end{minipage}
\caption{
The mean historical win percentage and the mean network value assigned to
(a) train and (b) validation synopses tend to decrease as
the number of \latentstates{} given to $\sigma$ increases.
}
\label{fig:uncertainty-graphs}
\end{figure}

Table~\ref{tab:exploration} shows the results of a grid search over
exploration constants and two bandit algorithms.
UCB1 \citep{kuleshov2014algorithms}
(with policy priors), which is used on the last line of
Algorithm~\ref{alg:stochastic-bandit},
is compared with ``a variant of PUCT'' (aVoP)
\citep{silver2016mastering, yu2019elf, lee2019minigo},
another popular bandit algorithm.
This experiment used $\kappa = 20\%$ and $\phi = 20\%$.
Figure~\ref{fig:uncertainty-graphs} show that
\penumbra{}'s value head accounts for
the uncertainty of the underlying \infostate{}.

\section{Per-bit saliency}
\label{sec:saliency}

Saliency methods may be able to identify which of the
synopsis feature planes are most important and which are least important.
Gradients only provide local information, and some saliency methods
fail basic sanity checks \citep{adebayo2018sanity}.
Higher quality saliency information may be surfaced by
integrating gradients over gradually-varied inputs
\citep{sundararajan2017axiomatic, kapishnikov2019xrai}
and by smoothing gradients locally
\citep{smilkov2017smoothgrad}.
Those saliency methods are not directly applicable to
discrete inputs such as the synopses used in this work.
So, this paper introduces a saliency method that aggregates
gradient information across two separate dimensions:
training examples and iterations.
Per-batch saliency (PBS) averages the absolute value of gradients
over random batches of test examples throughout training.
Similarly, per-bit saliency (PbS) averages
the absolute value of gradients over bits (with specific values)
within batches of test examples throughout training.
Gradients were taken both with respect to the loss
and with respect to the action policy.

\begin{figure}[t]
  \centering
  \begin{minipage}{0.5\linewidth}
  \centering
  \includegraphics[width=1\linewidth]{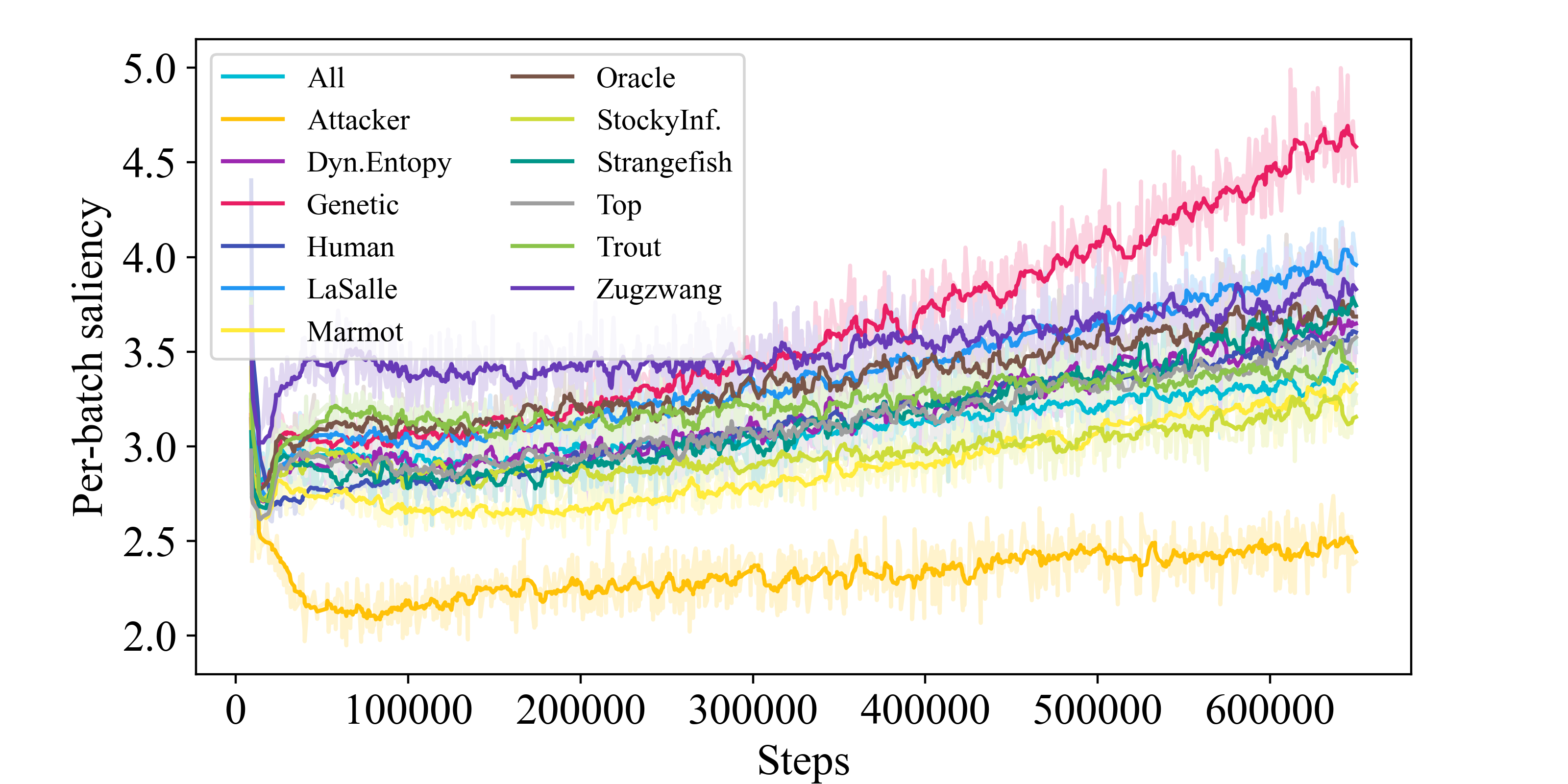}
  (a)
  \includegraphics[width=1\linewidth]{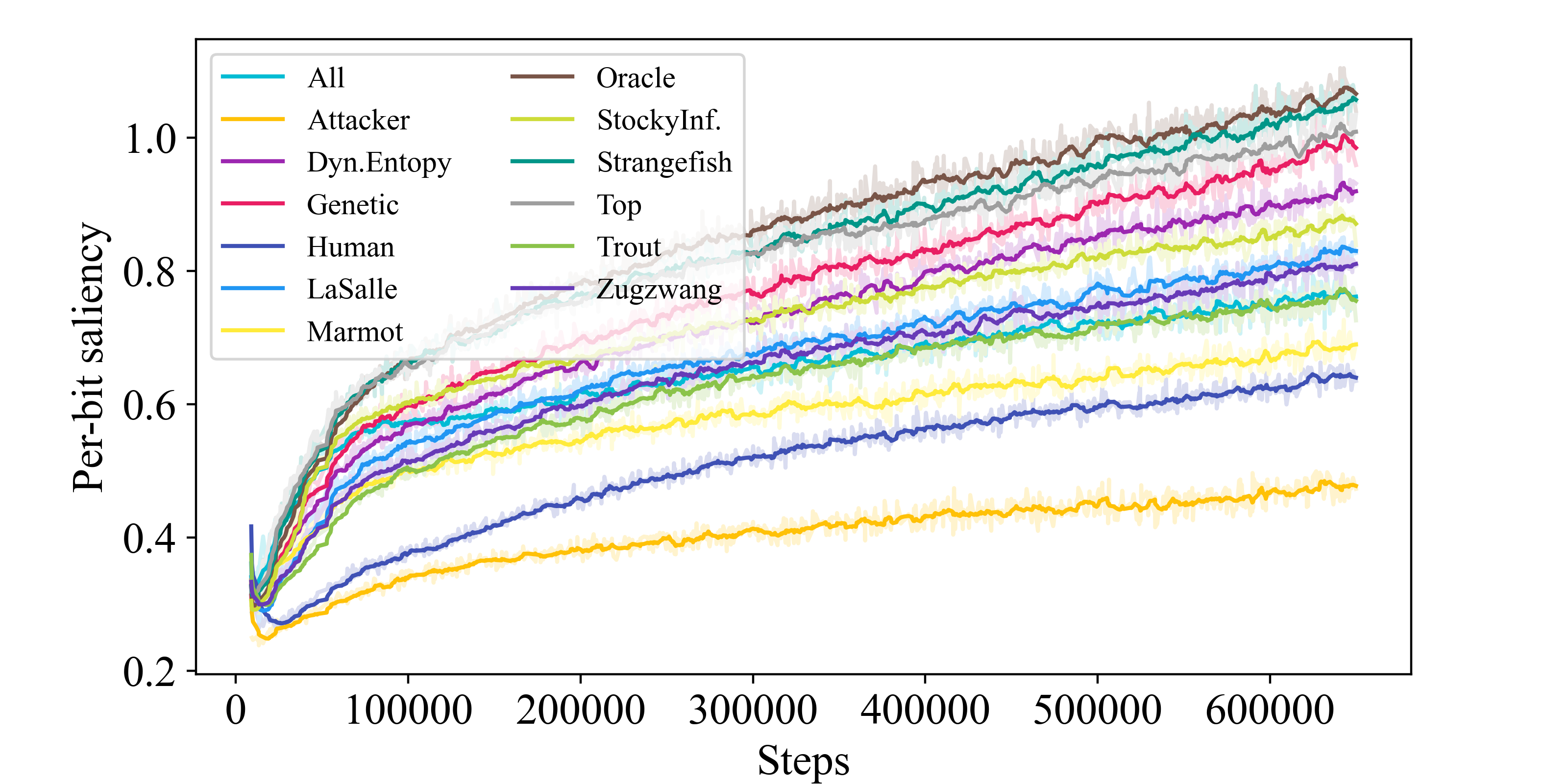}
  (b)
  \end{minipage}%
  \begin{minipage}{0.5\linewidth}
    \centering
  \includegraphics[width=1\linewidth]{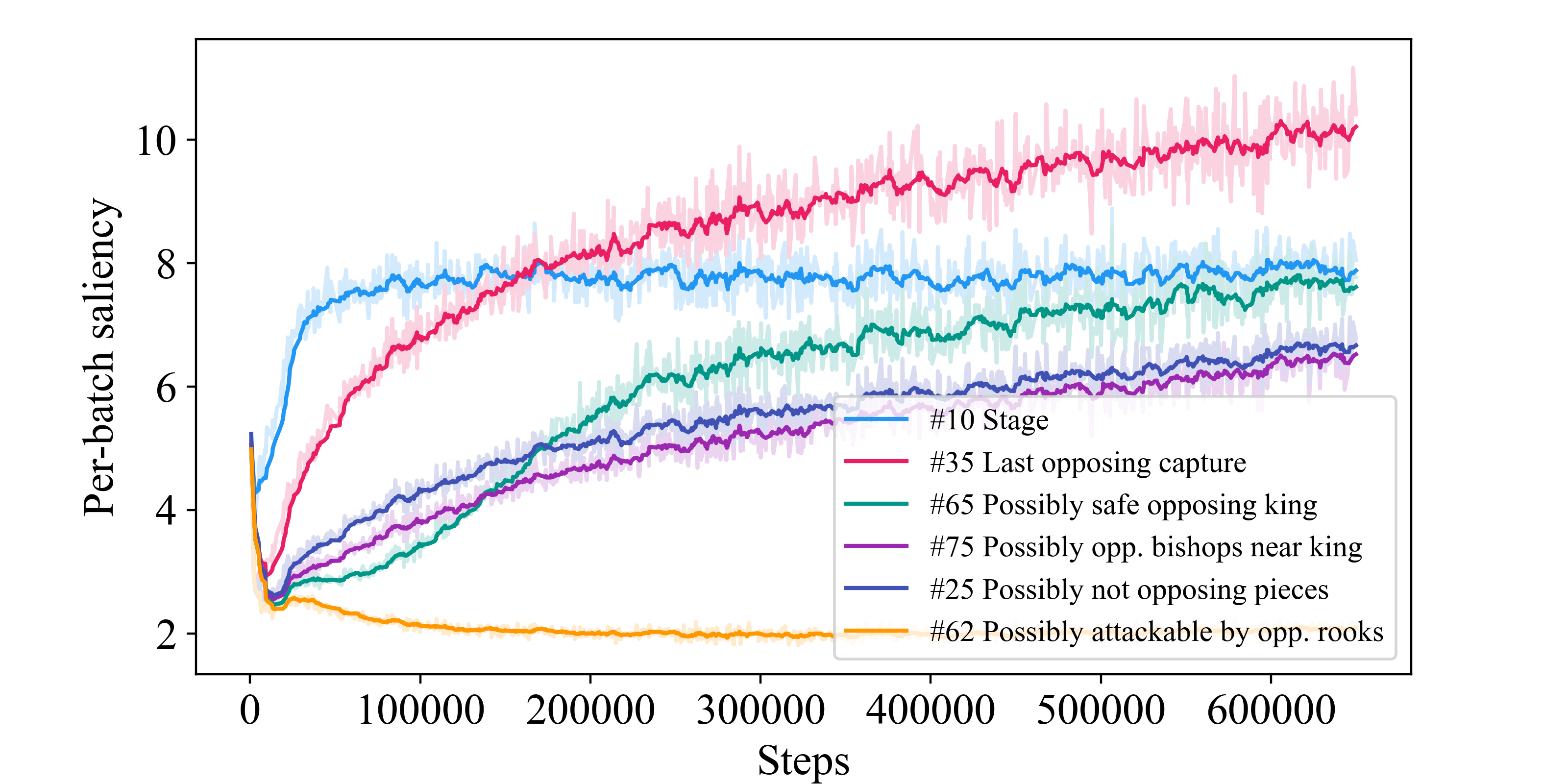}
  (c)
  \includegraphics[width=1\linewidth]{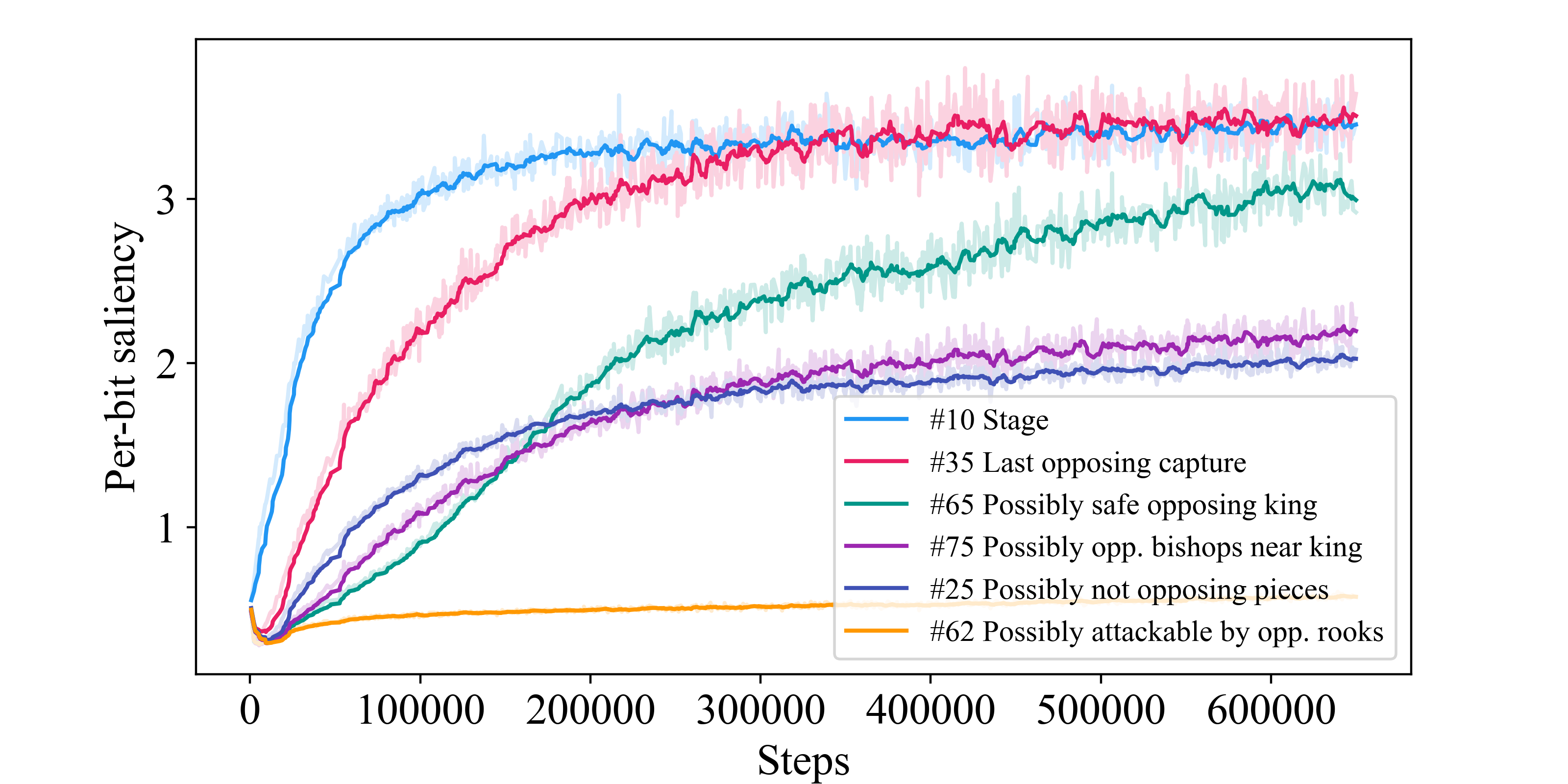}
  (d)
  \end{minipage}
\caption{
(a) The loss per-batch saliency (PBS) and
(b) the action per-bit saliency (PbS)
are taken on test examples during training.
These graphs show the saliency of
feature plane \#8, dark squares, for each headset in one training run.
The large gradients with respect to the loss
suggest that the \focusgenetic{} headset has overfit.
(c) The loss PBS and
(d) the action PbS
provide insight about which synopsis features are most useful.
The top-five most-salient feature planes
and the least-salient feature plane
for the \focustop{} headset from one training run are shown.
}
\label{fig:saliency-graphs}
\end{figure}

Figure~\ref{fig:saliency-graphs}
shows saliency information for input synopsis features used by \penumbra{}.
In order to validate that these saliency statistics are meaningful,
the model was retrained 104 times,
once with each feature removed \citep{hooker2018benchmark}.
Higher saliency is slightly correlated
with decreased performance when a feature is removed.
The correlation coefficient to the average change in accuracy is
$-0.208$ for loss-PBS, and $-0.206$ for action-PbS.
Explanations for the low correlation include
noise in the training process and
the presence of closely-related features.
Ultimately, the contribution of a feature during training is distinct from
how well the model can do without that feature.
Since some features are near-duplicates of others,
removing one may simply increase dependence on another.
Still, features with high saliency
--- such as the current stage (sense or move) and the location of the last capture ---
are likely to be the most important,
and features with low saliency may be considered for removal.
The appendix includes saliency statistics for each feature plane.

\section{Discussion}
\label{sec:discussion}

\paragraph{Broader impact}

DSMCP is more broadly applicable
than some prior algorithms for imperfect information games,
which are intractable in settings with large \infostates{}
and small amounts of shared knowledge \citep{brown2020combining}.
RBC and the related game Kriegspiel were motivated by
uncertainty in warfare \citep{newman2016reconnaissance}.
While playing board games is not dangerous in itself,
algorithms that account for uncertainty may become
effective and consequential in the real world.
In particular, since it focuses on exploiting weaknesses of other agents,
DSMCP could be applied in harmful ways.

\paragraph{Future work}

Planning in imperfect information games
is an active area of research \citep{russell2020artificial},
and RBC is a promising testing ground for such research.
\penumbra{} would likely benefit from further hyperparameter tuning
and potentially alternative corralled bandit algorithms \citep{arora2020corralling}.
Modeling an opponent poorly could be catastrophic;
algorithmic adjustments may lead to more-robust best-response strategies
\citep{ponsen2011acm}.
How much is lost by collapsing \infostates{} with synopses is unclear
and deserves further investigation.
Finally, the ``bitter lesson'' of machine learning \citep{sutton2019bitter}
suggests that a learned synopsis function may perform better.

\section*{Acknowledgements}

Thanks to the Johns Hopkins University Applied Physics Laboratory
for inventing such an intriguing game and for hosting RBC competitions.
Thanks to Ryan Gardner for valuable correspondence.
Thanks to Rosanne Liu, Joel Veness, Marc Lanctot, Zhe Zhao, and Zach Nussbaum
for providing feedback on early drafts.
Thanks to William Bernardoni for open sourcing high-quality baseline bots.
Thanks to Solidmind for the song ``Penumbra'',
which is an excellent soundtrack for programming.

\bibliography{main}
\bibliographystyle{plainnat}

\appendix

\input{appendix.tex}

\end{document}

%% file: setup.tex
\newcommand{\penumbra}{\texttt{Penumbra}}
\newcommand{\focusall}{\texttt{All}}
\newcommand{\focustop}{\texttt{Top}}
\newcommand{\focusstrangefish}{\texttt{StrangeFish}}
\newcommand{\focuslasalle}{\texttt{LaSalle}}
\newcommand{\focusdynentropy}{\texttt{Dyn.Entropy}}
\newcommand{\focusoracle}{\texttt{Oracle}}
\newcommand{\focuswbernar}{\texttt{StockyInfe.}}  % Wbernar
\newcommand{\focusstockyinference}{\texttt{StockyInference}}
\newcommand{\focusmarmot}{\texttt{Marmot}}
\newcommand{\focusgenetic}{\texttt{Genetic}}
\newcommand{\focuszugzwang}{\texttt{Zugzwang}}
\newcommand{\focustrout}{\texttt{Trout}}
\newcommand{\focushuman}{\texttt{Human}}
\newcommand{\focusattacker}{\texttt{Attacker}}
\newcommand{\focusrandom}{\texttt{Random}}

\newcommand{\tinyboard}{\includegraphics[width=0.25cm]{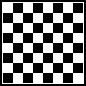}}

\definecolor{lightyellow}{rgb}{1,1,0.75}
\definecolor{lightgreen}{rgb}{0.875,1,0.875}
\definecolor{lightblue}{rgb}{0.875,0.875,1}
\definecolor{redred}{rgb}{1,0.75,0.75}

\newcommand{\playout}{playout}
\newcommand{\playouts}{playouts}
\newcommand{\Playouts}{Playouts}

\newcommand{\infostate}{infostate}
\newcommand{\infostates}{infostates}

\newcommand{\Infostates}{Infostates}
\newcommand{\longinfostate}{information state}
\newcommand{\longinfostates}{information states}

\newcommand{\latentstate}{world state}
\newcommand{\latentstates}{world states}

\newcommand{\featfunc}{g}
\newcommand{\numfeatures}{F}

\newcommand{\stimes}[0]{\small{\times}}
\newcommand{\spm}[0]{\small{\pm}}

\newcommand{\latstate}{x}
\newcommand{\latstates}{\mathcal{X}}
\newcommand{\latstateset}{X}
\newcommand{\actions}{\mathcal{A}}
\newcommand{\stateset}{X}
\newcommand{\beliefstate}{\hat{B}}
\newcommand{\prevsample}{J}
\newcommand{\oppsample}{J}
\newcommand{\sample}{I}
\newcommand{\observe}{\mathbf{o}}
\newcommand{\xa}{a(\latstate)}
\newcommand{\xtat}{a_t(\latstate_t)}
\newcommand{\counts}{\mathbf{N}}
\newcommand{\values}{\mathbf{Q}}

\newcommand{\actiontotalcount}{n}
\newcommand{\actioncounts}{\vec{n}}
\newcommand{\actionvalues}{\vec{q}}
\newcommand{\actionminimums}{\vec{m}}

\renewcommand{\algorithmiccomment}[1]{\textit{// #1}}

\renewcommand{\algorithmicrequire}{\textbf{Input:}}
\renewcommand{\algorithmicensure}{\textbf{Output:}}

\DeclareMathOperator*{\argmax}{argmax}

\newcommand{\Size}{2.5cm}% Adjust size of square as desired

\tikzset{Square/.style={
    inner sep=0pt,
    text width=\Size, 
    minimum size=\Size,
    draw=black,
    fill=yellow!20,
    align=center
    }
}

\newcommand{\featuretableheader}{\textbf{\#} &
\textbf{Description} &
\makecell{\textbf{Loss} \\ \textbf{PBS}} &
\makecell{\textbf{Action} \\ \textbf{PbS}} &
\makecell{\textbf{Action} \\ \textbf{PbS 0}} &
\makecell{\textbf{Action} \\ \textbf{PbS 1}} &
\makecell{\textbf{Action} \\ \textbf{PbS \#}} &
\makecell{\textbf{Ablation \%} \\ \textbf{Acc. Diff.} }\\
\hline}

\newcommand{\crossevaltableheader}{
 & \multicolumn{14}{c}{Dataset} \\
 \makecell{\\ \\ \\ \\ \\ Headset} &
\makecell{\rotatebox{90}{\focusall{}}} &
\makecell{\rotatebox{90}{\focustop{}}} &
\makecell{\rotatebox{90}{\focusstrangefish{}}} &
\makecell{\rotatebox{90}{\focuslasalle{}}} &
\makecell{\rotatebox{90}{\focusdynentropy{}}} &
\makecell{\rotatebox{90}{\focusoracle{}}} &
\makecell{\rotatebox{90}{\focuswbernar{}}} &
\makecell{\rotatebox{90}{\focusmarmot{}}} &
\makecell{\rotatebox{90}{\focusgenetic{}}} &
\makecell{\rotatebox{90}{\focuszugzwang{}}} &
\makecell{\rotatebox{90}{\focustrout{}}} &
\makecell{\rotatebox{90}{\focushuman{}}} &
\makecell{\rotatebox{90}{\focusattacker{}}} &
\makecell{\rotatebox{90}{\focusrandom{}}} \\
\hline}

%% file: first_figure.tex
\centering
\vspace{-25px}
\includegraphics[width=1\linewidth]{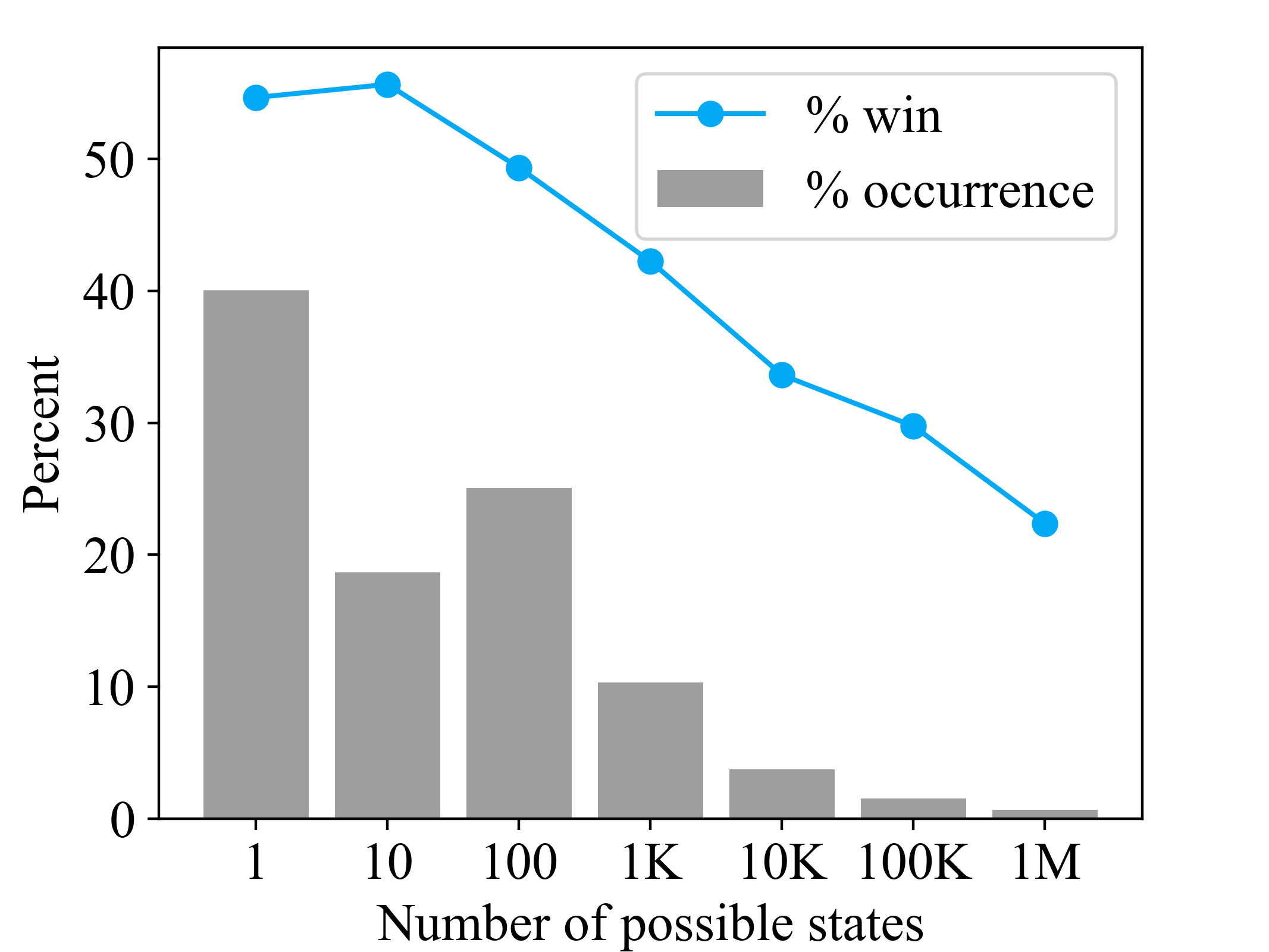}
\vspace{-8px}
\caption{
Within 98k historical reconnaissance blind chess (RBC)
games between non-random players,
the win percentage tends to decrease as the number of possible states increases.
The games were replayed while
tracking up to one million states.
Each bucket is labeled with an inclusive upper bound.
The median of the maximum number of
possible states encountered during a game is 4,869.
}
\label{fig:log-information-set-graph}

%% file: overview_figure.tex
\begin{tikzpicture}[
node distance=0.25cm and 0.25cm,
nodesty/.style={
  draw,
  text width=3.5cm,
  align=center,
  text height=0.65cm,
  text depth=0.1cm,
  rounded corners=2ex,
  thick,
},
popssty/.style={
  draw,
  text width=2.5cm,
  align=center,
  text height=0.32cm,
  text depth=0.1cm,
  thick,
},
circsty/.style={
  circle,
  draw=black,
  fill=white,
  align=center,
  thick,
  text width=0.9cm,
  text height=0.0cm,
  text depth=0.0cm,
  inner sep=0pt,
},
covesty/.style={
  circle,
  align=center,
  thick,
  text width=01cm,
  inner sep=0pt,
},
labelsty/.style={
}
]

\node[nodesty] (belief1) {  };
\node[nodesty, above=of belief1] (belief2) { $\qquad \qquad \qquad \qquad \dots$ };
\node[above=of belief2, draw=white, text height=0.04cm] (beliefdot) {$\dots$};
\node[nodesty, above=of beliefdot] (belief3) { $\qquad \qquad \qquad \qquad \dots$ };

\node[circsty, above=of belief3, xshift=1.75cm, yshift=-0.05cm] (shadow1) { };
\node[circsty, above=of belief3, xshift=1.85cm] (sitch1) { \vspace{-2pt}\tinyboard \, \tinyboard \\ \tinyboard };

\node[circsty, above=of sitch1, xshift=1.75cm, yshift=-0.15cm] (shadow2) { };
\node[circsty, above=of sitch1, xshift=1.85cm, yshift=-0.1cm] (sitch2) { \vspace{-4pt}\tinyboard \\ \vspace{0.5pt}\tinyboard \, \tinyboard };

\node[circsty, above=of sitch1, xshift=-1.85cm, yshift=-0.15cm] (shadow3) { };
\node[circsty, above=of sitch1, xshift=-1.75cm, yshift=-0.1cm] (sitch3) { \vspace{-5pt} \tinyboard \\ \vspace{1.5pt} \tinyboard \, \tinyboard };

\node[circsty, above=of sitch1, xshift=0.25cm, yshift=0.15cm] (shadow4) { };
\node[circsty, above=of sitch1, xshift=0.35cm, yshift=0.2cm] (sitch4) {\tinyboard \, \tinyboard };

\node[above=of sitch2] (cont1) {$\dots$};
\node[above=of sitch3] (cont2) {$\dots$};
\node[above=of sitch4] (cont3) {$\dots$};

\node[labelsty, left=-0.75cm of belief1] (label1) { $\beliefstate_0$ };
\node[labelsty, left=-0.75cm of belief2] (label2) { $\beliefstate_1$ };
\node[labelsty, left=-0.75cm of belief3] (label3) { $\beliefstate_t$ };

\node[popssty, right=of belief1] (popset1) { \tinyboard };
\node[popssty, right=of belief2] (popset2) {  \, \tinyboard \, \tinyboard \, $\dots$ };
\node[above=of popset2, draw=white, text height=0.04cm] (popsetdot) {$\dots$};
\node[popssty, right=of belief3] (popset3) {  \, \tinyboard \, \tinyboard \, $\dots$ };

\node[labelsty, left=-0.75cm of popset1] (label4) { $\stateset_0$ };
\node[labelsty, left=-0.75cm of popset2] (label5) { $\stateset_1$ };
\node[labelsty, left=-0.75cm of popset3] (label6) { $\stateset_t$ };

\node[circsty, above=of belief1, yshift=-1.215cm] (bsitch1) {\tinyboard};

\node[circsty, above=of belief2, yshift=-1.215cm, xshift=-0.5cm] (bsitch2) {\vspace{-3pt}\tinyboard \, \tinyboard \\ \tinyboard};
\node[circsty, above=of belief2, yshift=-1.215cm, xshift=0.5cm] (bsitch3) {\tinyboard \, \tinyboard};

\node[circsty, above=of belief3, yshift=-1.215cm, xshift=-0.5cm] (bsitch2) {\vspace{-4pt}\tinyboard \, \tinyboard \\ \vspace{1.5pt}\tinyboard \hspace{1pt}\tinyboard};
\node[circsty, above=of belief3, yshift=-1.215cm, xshift=0.5cm] (bsitch3) {\vspace{-4pt} \tinyboard \\ \vspace{1.5pt}\tinyboard};

\draw[->, thick] (belief1) -- (belief2);
\draw[->, thick] (belief2) -- (beliefdot);
\draw[->, thick] (beliefdot) -- (belief3);

\draw[->, thick] (popset1) -- (popset2);
\draw[->, thick] (popset2) -- (popsetdot);
\draw[->, thick] (popsetdot) -- (popset3);

\draw[->, thick] (belief3) -- (shadow1);
\draw[->, thick] (popset3) -- (sitch1);
\draw[->, thick] (sitch1) -- (sitch2);
\draw[->, thick] (sitch1) -- (sitch3);
\draw[->, thick] (sitch1) -- (sitch4);

\draw[->, thick] (sitch2) -- (cont1);
\draw[->, thick] (sitch3) -- (cont2);
\draw[->, thick] (sitch4) -- (cont3);

\end{tikzpicture}

%% file: algorithms_one.tex
\begin{algorithm}[H]
  \caption{\texttt{Bandit} -- Action selection with a stochastic multi-armed bandit}
  \label{alg:stochastic-bandit}
\begin{algorithmic} % [1]

  \STATE {\bfseries Given:} $c > 0, m \ge 0$ 
  \STATE {\bfseries Input:} policy $\pi$,
                            visit counts $\actioncounts$, 
                            value totals $\actionvalues$ 
  
  \STATE $\actiontotalcount \gets \sum_a {\actioncounts_a}$

  \IF{$e^{-mn} > \texttt{Uniform}([0, 1])$ and $\neg \text{root}$}
    \STATE {\bfseries return}
    $a \gets $ an action selected by policy $\pi$
  
  \ELSE
    
    \STATE {\bfseries return}
    $a \gets \argmax_a \left(\frac{ \actionvalues_a }{ \actioncounts_a }
             + c \pi_a \sqrt{\frac{\ln{\actiontotalcount}}{\actioncounts_a}} \right)$

  \ENDIF

\end{algorithmic}
\end{algorithm}

\begin{algorithm}[H]
  \caption{\texttt{DrawSample} -- Sample selection with rejection}
  \label{alg:prepare-sample}
\begin{algorithmic}
  \STATE {\bfseries Given:} $k, \ell \in \mathbb{Z}^+$, synopsis function $\sigma$,
  \STATE \phantom{{\bfseries Given:}} policy $\hat{\tau}$
  \STATE {\bfseries Input:} previous belief distribution $\beliefstate$, possible
  \STATE \phantom{{\bfseries Input:}} states $\stateset$, visit counts $\counts$, value totals $\values$
  
  \FOR{$0$ {\bfseries to} $k$}
  
    \STATE $\prevsample \gets$ random sample from $\beliefstate$
    \STATE $a \gets\texttt{Bandit}(\hat{\tau}(\sigma(J)), \counts_{J}, \values_{J})$
    
    \STATE $\sample \gets \{\xa : \latstate \in \prevsample \}$
    
    \STATE $\sample \gets$ random
           % subset of
           $\ell$ states from $I$ {\bfseries if} $|\sample| > \ell$
    
    \STATE {\bfseries return} $\sample$ {\bfseries if} $\sample \cap \stateset \neq \varnothing$
  \ENDFOR
  \STATE {\bfseries return} $\{$ random state from $\stateset$ $\}$
\end{algorithmic}
\end{algorithm}

%% file: algorithms.tex
\begin{minipage}[t]{.5\linewidth}
\begin{algorithm}[H]
  \caption{\texttt{ChooseAction} -- UCT \playouts}
  \label{alg:choose-action}
\begin{algorithmic} % [1]
  \STATE {\bfseries Given:} $b, d, \ell, n_{\text{vl}}, z \in \mathbb{Z}^+$,
                            synopsis func. $\sigma$,
  \STATE \phantom{{\bfseries Given:}} 
                            policy $\pi$, opp. policy $\hat{\tau}$, value func. $\nu$
  \STATE {\bfseries Input:} belief distribution $\beliefstate$, possible states $\stateset$,
  \STATE \phantom{{\bfseries Input:}} visit counts $\counts$, value totals $\values$
  
  \STATE $\bar{\pi} \gets$ average of $\pi \circ \sigma$ over $b$ samples from $\beliefstate$
  
  \WHILE{time is left}
    \STATE $\oppsample \gets$ random sample from $\beliefstate$
    \STATE $\sample \gets$ random $\ell$ states in $\stateset$ including one in $\oppsample$

    \STATE $a_0 \gets \texttt{Bandit}(\bar{\pi},
                                      \counts_{\text{root}},
                                      \values_{\text{root}}
                                      )$
    
    \FOR{$t = 0$ {\bfseries to} $d$}
      
      \STATE $\latstate{}_{t} \gets$ the one state in $I \cap J$
      
      \STATE $\observe \gets$ observations when $a_{t}$ is played on $\latstate{}_{t}$
      
      \STATE $\sample \gets \{\xa : \latstate \in \sample, a \in \actions_\sample$,
                            % $ \mathbf{o}_{\text{self}}(xa) = 1 \}$
                            $ \observe_{\text{self}} \sim \xa\}$
      \STATE $\oppsample \gets \{\xa : \latstate \in \oppsample, a \in \actions_\oppsample$,
                               % $\mathbf{o}_{\text{opp.}}(xa) = 1 \}$
                               $\observe_{\text{opp}} {\sim} \, \xa\}$
      
      \STATE $\sample \gets$ random $\ell$ states in $\sample$ including $\xtat$
      \STATE $\oppsample \gets$ random $\ell$ states in $\oppsample$ including $\xtat$
      
      \STATE $K_t \gets I$ and $\mu \gets \pi$ {\bfseries if} $I$ is to act
      \STATE $K_t \gets J$ and $\mu \gets \hat{\tau}$ {\bfseries if} $J$ is to act
      
      \STATE $d  \mathrel{+}= 1$ {\bfseries if} $\counts_{K_t, a_t} > z$
      
      \FOR{$i = 0$ {\bfseries to} $t$}
        \STATE $\counts_{K_i, a_{i}} \mathrel{+}= n_{\text{vl}}$
      \ENDFOR
      
      \STATE $a_{t+1} \gets \texttt{Bandit}(\mu(\sigma(K_t)), \counts_{K_t}, \values_{K_t})$
      \STATE $q \gets \nu(\sigma(K_t))$
      
      \FOR{$i = 0$ {\bfseries to} $t$}
        \STATE $\values_{K_i, a_i} \mathrel{+}= q$
                                    {\bfseries if} $\rho(K_t) {=} \rho(K_i)$
                                    {\bfseries else} $-q$
        \STATE $\counts_{K_i, a_i} \mathrel{+}= 1 - n_{\text{vl}}$
      \ENDFOR
    \ENDFOR
    
  \ENDWHILE
  
  \STATE {\bfseries return} $a \gets \argmax_{a} \values_{\text{root}}$

\end{algorithmic}
\end{algorithm}
\end{minipage}%
\begin{minipage}[t]{.5\linewidth}
\begin{algorithm}[H]
  \caption{\texttt{PlayGame} -- DSMCP}
  \label{alg:play-game}
\begin{algorithmic}
  \STATE {\bfseries Given:} $n_{\text{particles}} \in \mathbb{Z}^+$
  
  \STATE $\counts_{*,*} \gets \mathbf{0}$
    \COMMENT{Visits $\forall$ (sample, action) $\in \mathcal{L} \times \actions$}
  \STATE $\values_{*,*} \gets \mathbf{0}$
    \COMMENT{Values $\forall$ (sample, action) $\in \mathcal{L} \times \actions$}
  
  \STATE $\stateset_0 \gets \{\latstate{}_0\}$
  \STATE $\beliefstate_0 \gets \{\stateset_0\}$

  \WHILE{the game is not over}
    \STATE $t \gets \text{current turn}$
    \STATE $\observe \gets \text{current observation}$
    
    \STATE \COMMENT{Track all possible world states}
    \STATE $\stateset_t \hspace{-2pt}\gets\hspace{-3pt} \{\xa : \latstate\hspace{-1pt} \in\hspace{-2pt} \stateset_{t-1}, a\hspace{-1pt} \in\hspace{-1pt} \actions_{\latstate}$,
                                     $\observe_{\text{self}} \hspace{-2pt}\sim\hspace{-2pt} \xa\}$
    
    \STATE \COMMENT{Filter belief states with the new information}
    \FOR{$i = t - 1$ {\bfseries to} $0$}
      \STATE $\stateset_i \gets \{\latstate \in \stateset_i : \exists a \in \actions_{x}$
      \STATE \phantom{$\stateset_i \gets \{\latstate \in \stateset_i : $}
          such that $\xa \in \stateset_{i+1} \}$
      \STATE $\beliefstate_i \gets \{\sample \in \beliefstate_i :
                                     \sample \cap \stateset_i \neq \varnothing \}$
    \ENDFOR
    
    \STATE \COMMENT{Repopulate belief states with new particles}
    \WHILE{opponent to act or $|\beliefstate_t| < n_{\text{particles}}$}

      \STATE $i \gets$ smallest $i > 0$ s.t. $|\beliefstate_i| < n_{\text{particles}}$
      
      \STATE $\sample \gets \texttt{DrawSample}(\beliefstate_{i-1}, \stateset_i, \counts, \values)$
      \STATE $\beliefstate_i \gets \beliefstate_i \cup \{\sample\}$
    \ENDWHILE
    \IF{self to act}
      \STATE $a \gets \texttt{ChooseAction}(\beliefstate_t, \stateset_t, \counts, \values)$ 
      \STATE Perform action $a$

    \ENDIF{}
  \ENDWHILE
\end{algorithmic}
\end{algorithm}
\end{minipage}

%% file: example_game.tex
\begin{figure}[t]
\begin{center}
\begin{minipage}{0.5\columnwidth}
    \centering
    \includegraphics[width=0.5\linewidth]{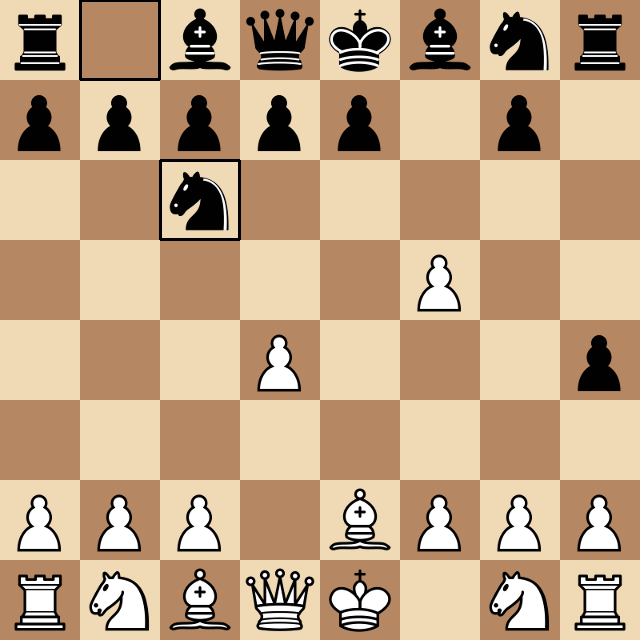}
\end{minipage}%
\begin{minipage}{0.5\columnwidth}
    \centering
    \rowcolors{2}{gray!10}{white}
    \begin{tabular}{|c|c|c|c|c|}
        \hline
        \textbf{\#} & \textbf{White} & \textbf{Black} \\
        \hline
        1 & \texttt{g6} : \texttt{e2e4} & \texttt{e3} : \texttt{h7h5} \\
        2 & \texttt{g7} : \texttt{d2d4} & \texttt{f2} : \texttt{f7f5} \\
        3 & \texttt{g6} : \texttt{e4f5} & \texttt{e4} : \texttt{h5h4} \\
        4 & \texttt{d7} : \texttt{f1e2} & \cellcolor{blue!16}\texttt{g4} : \texttt{b8c6} \\
        5 & \texttt{g7} : \texttt{e2h5} & \texttt{g4} : \texttt{h8h5} \\
        6 & \texttt{b7} : \texttt{d1h5} & \texttt{g5} : \texttt{g7g6} \\
        7 & \texttt{e7} : \texttt{h5e8} & \texttt{g6} : \texttt{d7d6} \\
        8 & \texttt{g6} : \texttt{g6e8} & \\
        \hline
    \end{tabular}
\end{minipage}%
\end{center}
\caption{
\penumbra{} played as White in this short game.
In the position shown on the left,
Black just moved a knight from \texttt{b8} to \texttt{c6}.
From White's perspective,
the black pieces could
be placed in 238 different ways.
Figure~\ref{fig:sample-situation} shows
a set of synopsis bitboards
for this \longinfostate{}.
}
\label{fig:sample-game}
\end{figure}

%% file: bitboards_wide.tex
\newcommand{\bitboard}[2]{%
  \makecell{\includegraphics[width=0.0407\linewidth]{images/situation1/feature_#1.png} \vspace{-4px} \\ \small{#2}}}

\setlength{\tabcolsep}{0.2pt}
\begin{tabular}{ccccccccccccccccccccc}
    \bitboard{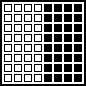}{0} &
    \bitboard{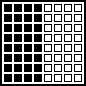}{1} &
    \bitboard{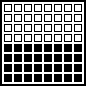}{2} &
    \bitboard{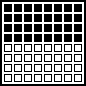}{3} &
    \bitboard{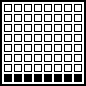}{4} &
    \bitboard{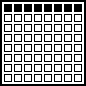}{5} &
    \bitboard{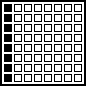}{6} &
    \bitboard{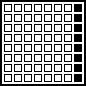}{7} &
    \bitboard{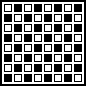}{8} &
    \bitboard{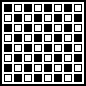}{9} &
    \bitboard{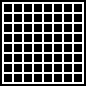}{10} &
    \bitboard{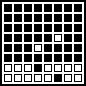}{11} &
    \bitboard{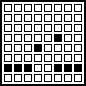}{12} &
    \bitboard{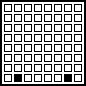}{13} &
    \bitboard{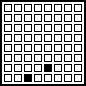}{14} &
    \bitboard{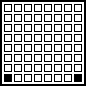}{15} &
    \bitboard{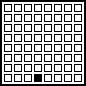}{16} &
    \bitboard{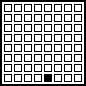}{17} &
    \bitboard{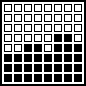}{18} &
    \bitboard{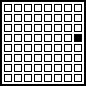}{19} &
    \bitboard{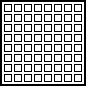}{20} \\
    \bitboard{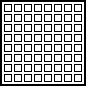}{21} &
    \bitboard{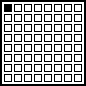}{22} &
    \bitboard{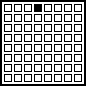}{23} &
    \bitboard{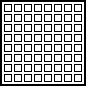}{24} &
    \bitboard{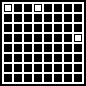}{25} &
    \bitboard{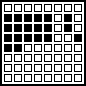}{26} &
    \bitboard{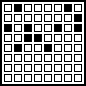}{27} &
    \bitboard{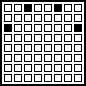}{28} &
    \bitboard{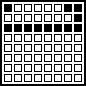}{29} &
    \bitboard{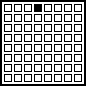}{30} &
    \bitboard{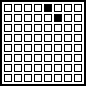}{31} &
    \bitboard{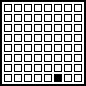}{32} &
    \bitboard{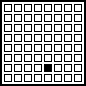}{33} &
    \bitboard{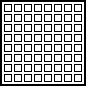}{34} &
    \bitboard{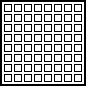}{35} &
    \bitboard{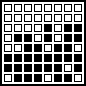}{36} &
    \bitboard{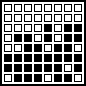}{37} &
    \bitboard{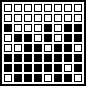}{38} &
    \bitboard{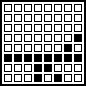}{39} &
    \bitboard{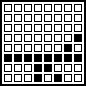}{40} &
    \bitboard{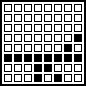}{41} \\
    \bitboard{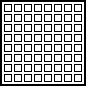}{42} &
    \bitboard{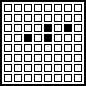}{43} &
    \bitboard{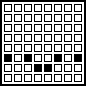}{44} &
    \bitboard{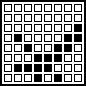}{45} &
    \bitboard{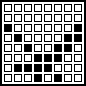}{46} &
    \bitboard{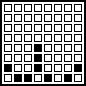}{47} &
    \bitboard{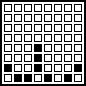}{48} &
    \bitboard{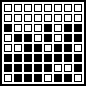}{49} &
    \bitboard{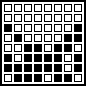}{50} &
    \bitboard{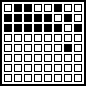}{51} &
    \bitboard{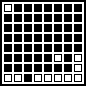}{52} &
    \bitboard{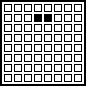}{53} &
    \bitboard{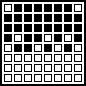}{54} &
    \bitboard{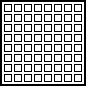}{55} &
    \bitboard{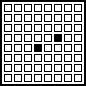}{56} &
    \bitboard{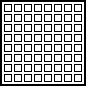}{57} &
    \bitboard{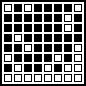}{58} &
    \bitboard{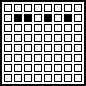}{59} &
    \bitboard{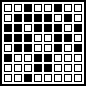}{60} &
    \bitboard{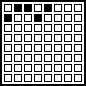}{61} &
    \bitboard{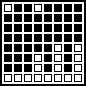}{62} \\
    \bitboard{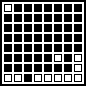}{63} &
    \bitboard{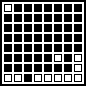}{64} &
    \bitboard{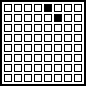}{65} &
    \bitboard{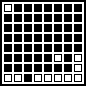}{66} &
    \bitboard{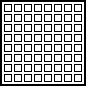}{67} &
    \bitboard{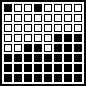}{68} &
    \bitboard{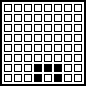}{69} &
    \bitboard{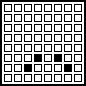}{70} &
    \bitboard{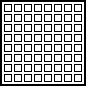}{71} &
    \bitboard{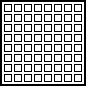}{72} &
    \bitboard{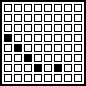}{73} &
    \bitboard{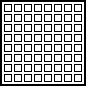}{74} &
    \bitboard{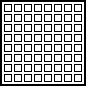}{75} &
    \bitboard{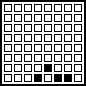}{76} &
    \bitboard{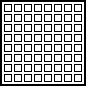}{77} &
    \bitboard{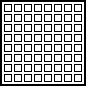}{78} &
    \bitboard{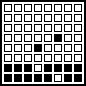}{79} &
    \bitboard{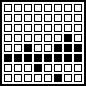}{80} &
    \bitboard{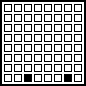}{81} &
    \bitboard{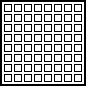}{82} &
    \bitboard{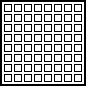}{83} \\
    \bitboard{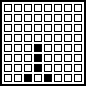}{84} &
    \bitboard{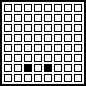}{85} &
    \bitboard{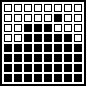}{86} &
    \bitboard{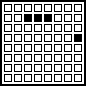}{87} &
    \bitboard{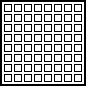}{88} &
    \bitboard{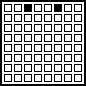}{89} &
    \bitboard{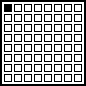}{90} &
    \bitboard{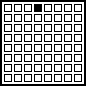}{91} &
    \bitboard{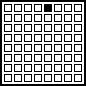}{92} &
    \bitboard{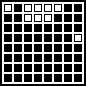}{93} &
    \bitboard{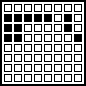}{94} &
    \bitboard{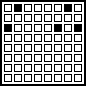}{95} &
    \bitboard{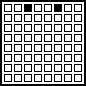}{96} &
    \bitboard{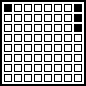}{97} &
    \bitboard{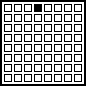}{98} &
    \bitboard{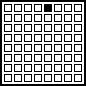}{99} &
    \bitboard{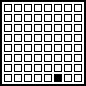}{100} &
    \bitboard{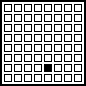}{101} &
    \bitboard{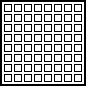}{102} &
    \bitboard{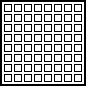}{103} &
\end{tabular}

%% file: architecture.tex
\begin{tikzpicture}[
node distance=0.25cm and 0.25cm,
nodesty/.style={
  draw,
  text width=1.63cm,
  align=center,
  text height=2ex,
  text depth=0.1ex,
  thick,
},
title/.style={
  text width=12.75cm,
  font=\color{black!50},
},
typetag/.style={
  rectangle,
  draw=black!50,
  font=\scriptsize
}
]
\node[nodesty, text width=4.16cm] (input) {$8 \stimes 8 \stimes 104$ input};
\node[nodesty, text width=5.12cm, below=of input, double, double distance=0.23mm] (block0) {$8 \stimes 8 \stimes 128$ res. block};
\node[nodesty,text width=5cm, below=of block0, draw=white] (blockdot) {... 8 residual blocks omitted ...};
\node[nodesty,text width=5.12cm, below=of blockdot, double, double distance=0.23mm] (block9) {$8 \stimes 8 \stimes 128$ res. block};

\node[title, below=of block9] (headset0) { };
\node[title, below=of block9, yshift=-0.5mm, xshift=-0.5mm] (headset1) { };
\node[title, below=of block9, yshift=-1mm, xshift=-1mm] (headset2) { };
\node[title, below=of block9, yshift=-1.5mm, xshift=-1.5mm] (headset3) { };
\node[title, below=of block9, yshift=-2mm, xshift=-2mm] (headset4) { };
\node[title, below=of block9, yshift=-2.5mm, xshift=-2.5mm] (headset5) { };
\node[title, below=of block9, yshift=-3mm, xshift=-3mm] (headset6) { };
\node[title, below=of block9, yshift=-3.5mm, xshift=-3.5mm] (headset7) { };
\node[title, below=of block9, yshift=-4mm, xshift=-4mm] (headset8) { };
\node[title, below=of block9, yshift=-4.5mm, xshift=-4.5mm] (headset9) { };
\node[title, below=of block9, yshift=-5mm, xshift=-5mm] (headset10) { };
\node[title, below=of block9, yshift=-5.5mm, xshift=-5.5mm] (headset11) { };
\node[title, below=of block9, yshift=-6mm, xshift=-6mm] (headset12) { };
\node[title, below=of block9, yshift=-6.5mm, xshift=-6.5mm] (headset13) { };

\draw [draw=black!50, fill=white] (headset0.north west) rectangle ($(headset0.north east) - (0, 3cm)$);

\draw [draw=black!50, fill=white] (headset1.north west) rectangle ($(headset1.north east) - (0, 3cm)$);
\draw [draw=black!50, fill=white] (headset2.north west) rectangle ($(headset2.north east) - (0, 3cm)$);
\draw [draw=black!50, fill=white] (headset3.north west) rectangle ($(headset3.north east) - (0, 3cm)$);
\draw [draw=black!50, fill=white] (headset4.north west) rectangle ($(headset4.north east) - (0, 3cm)$);
\draw [draw=black!50, fill=white] (headset5.north west) rectangle ($(headset5.north east) - (0, 3cm)$);
\draw [draw=black!50, fill=white] (headset6.north west) rectangle ($(headset6.north east) - (0, 3cm)$);
\draw [draw=black!50, fill=white] (headset7.north west) rectangle ($(headset7.north east) - (0, 3cm)$);
\draw [draw=black!50, fill=white] (headset8.north west) rectangle ($(headset8.north east) - (0, 3cm)$);
\draw [draw=black!50, fill=white] (headset9.north west) rectangle ($(headset9.north east) - (0, 3cm)$);
\draw [draw=black!50, fill=white] (headset10.north west) rectangle ($(headset10.north east) - (0, 3cm)$);
\draw [draw=black!50, fill=white] (headset11.north west) rectangle ($(headset11.north east) - (0, 3cm)$);
\draw [draw=black!50, fill=white] (headset12.north west) rectangle ($(headset12.north east) - (0, 3cm)$);

\draw [draw=black!50, fill=white] (headset13.north west) rectangle ($(headset13.north east) - (0, 3cm)$);

\node[title, below=of block9, yshift=-6.5mm, xshift=-6.5mm] (headset) { Headset };

\node[nodesty, below=of block9, yshift=-1.3cm, xshift=-1.7cm] (value) { $8 \stimes 8 \stimes 8$ };
\node[nodesty, below=of block9, left=of value] (pass) { $8 \stimes 8 \stimes 4$ };
\node[nodesty, below=of block9, left=of pass, double, double distance=0.23mm] (policy) { $8 \stimes 8 \stimes 128$ };
\node[nodesty, below=of block9, right=of value] (soonwin) { $8 \stimes 8 \stimes 4$ };
\node[nodesty, below=of block9, right=of soonwin] (soonlose) { $8 \stimes 8 \stimes 4$ };
\node[nodesty, below=of block9, right=of soonlose] (piececount) { $8 \stimes 8 \stimes 216$ };

\draw[->, thick] (input) -- (block0);
\draw[->, thick] (block0) -- (blockdot);
\draw[->, thick] (blockdot) -- (block9);
\draw[->, thick] (block9.south) .. controls +(down:1cm) and +(up:1cm) .. (value.north);
\draw[->, thick] (block9.south) .. controls +(down:1cm) and +(up:1cm) .. (policy.north);
\draw[->, thick] (block9.south) .. controls +(down:1cm) and +(up:1cm) .. (pass.north);
\draw[->, thick] (block9.south) .. controls +(down:1cm) and +(up:1cm) .. (piececount.north);
\draw[->, thick] (block9.south) .. controls +(down:1cm) and +(up:1cm) .. (soonwin.north);
\draw[->, thick] (block9.south) .. controls +(down:1cm) and +(up:1cm) .. (soonlose.north);

\node[nodesty, below=of policy] (policy1) { $8 \stimes 8 \stimes 65 $ };
\draw[->, thick] (policy.south) -- (policy1.north);
\node[nodesty, below=of policy1] (policy2) { Policy };
\draw[->, thick] (policy1.south) -- (policy2.north);

\node[nodesty, below=of pass] (pass1) { 64 dense };
\draw[->, thick] (pass.south) -- (pass1.north);
\node[nodesty, draw, align=center, thick, right=of policy2, xshift=-0.275cm, text width=0.63cm,] (pass2) {
Pass
};
\draw[->, thick] (pass1.south) .. controls +(down:0.3cm) and +(right:0.3cm) .. (pass2.east);

\node[nodesty, below=of value] (value1) { 256 dense };
\draw[->, thick] (value.south) -- (value1.north);
\node[nodesty, below=of value1] (value2) { Value };
\draw[->, thick] (value1.south) -- (value2.north);

\node[nodesty, below=of soonwin] (soonwin1) { 256 dense };
\draw[->, thick] (soonwin.south) -- (soonwin1.north);
\node[nodesty, below=of soonwin1] (soonwin2) { SoonWin };
\draw[->, thick] (soonwin1.south) -- (soonwin2.north);

\node[nodesty, below=of soonlose] (soonlose1) { 256 dense };
\draw[->, thick] (soonlose.south) -- (soonlose1.north);
\node[nodesty, below=of soonlose1] (soonlose2) { SoonLose };
\draw[->, thick] (soonlose1.south) -- (soonlose2.north);

\node[nodesty, below=of piececount] (piececount1) { $8 \stimes 8 \stimes 216$ };
\draw[->, thick] (piececount.south) -- (piececount1.north);
\node[nodesty, below=of piececount1] (piececount2) { PieceCount };
\draw[->, thick] (piececount1.south) -- (piececount2.north);

\end{tikzpicture}

%% file: headset_summary.tex
\begin{table*}[t]
\caption{Summary of headset training data and resulting validation accuracies}
\label{tab:headset-descriptions}
\vskip 0.1in
\begin{center}
\begin{tabular}{rrrrcccrrrr}
     \rotatebox{90}{\textbf{Name}} &
     \rotatebox{90}{\textbf{\# Games}} &
     \rotatebox{90}{\makecell[l]{\textbf{\# Train}\\\textbf{Actions}}} &
     \rotatebox{90}{\makecell[l]{\textbf{\# Validation}\\\textbf{Actions}}} &
     \rotatebox{90}{\makecell[l]{\textbf{Multiplicity}}} &
     \rotatebox{90}{\makecell[l]{\textbf{Training}\\\textbf{Weight}}} & 
     \rotatebox{90}{\makecell[l]{\textbf{Stop}\\\textbf{Gradient}}} &
     \rotatebox{90}{\makecell[l]{\textbf{\% Top-1}\\\textbf{Accuracy}}} &
     \rotatebox{90}{\makecell[l]{\textbf{\% Top-5}\\\textbf{Accuracy}}} &
     \rotatebox{90}{\makecell[l]{\textbf{\% Winner}\\\textbf{Accuracy}}} &
     \\
     \hline
     \focusall{}         & 85.6k & 16.8M & 1.8M & \phantom{0}4 &           15 & No  & 33.6 & 62.0 & 74.8 \\% & 79.06 \\
     \focustop{}         & 49.5k & 16.0M & 1.6M & \phantom{0}$\sim$9.6 &   13 & No  & 43.1 & 73.4 & 82.4 \\% & 75.90 \\
     \focusstrangefish{} & 13.3k &  8.1M & 0.8M & $\sim$20.7   &           10 & No  & 45.9 & 74.3 & 86.6 \\% & 77.06 \\
     \focuslasalle{}     &  3.7k &  4.3M & 0.4M & 32           & \phantom{0}4 & No  & 36.4 & 68.2 & 69.7 \\% & 71.93 \\
     \focusdynentropy{}  &  7.8k &  7.8M & 0.8M & 32           & \phantom{0}7 & No  & 50.0 & 76.7 & 78.9 \\% & 76.16 \\
     \focusoracle{}      & 20.3k & 17.6M & 1.9M & 32           & \phantom{0}9 & No  & 49.3 & 77.4 & 81.4 \\% & 76.78 \\
     \focuswbernar{}     & 10.7k & 13.5M & 1.4M & 16           & \phantom{0}9 & No  & 45.2 & 73.6 & 68.3 \\% & 72.07 \\
     \focusmarmot{}      & 10.2k &  3.9M & 0.4M & 16           & \phantom{0}8 & No  & 24.6 & 55.4 & 80.8 \\% & 80.66 \\
     \focusgenetic{}     &  5.6k &  2.3M & 0.2M & 16           & \phantom{0}8 & No  & 40.4 & 70.5 & 77.9 \\% & 78.68 \\
     \focuszugzwang{}    & 10.9k &  3.1M & 0.3M & $\sim$12.0   & \phantom{0}5 & No  & 47.0 & 71.6 & 80.1 \\% & 78.21 \\
     \focustrout{}       & 18.0k &  5.1M & 0.5M & 16           & \phantom{0}5 & No  & 41.8 & 63.4 & 79.8 \\% & 82.81 \\
     \focushuman{}       &  6.3k &  1.5M & 0.2M & 16           & \phantom{0}2 & No  & 24.9 & 51.9 & 73.3 \\% & 75.28 \\
     \focusattacker{}    & 15.3k &  1.7M & 0.2M & 16           & \phantom{0}4 & No  & 45.1 & 61.9 & 80.1 \\% & 93.43 \\
     \focusrandom{}      & 17.0k &  0.6M &  76k & \phantom{0}2 & \phantom{0}1 & Yes &  4.5 & 20.7 & 94.7 \\% & 85.19 \\
\end{tabular}
\end{center}
\vskip -0.1in
\end{table*}

%% file: result_tables.tex
\begin{table}[t]
\begin{minipage}{.5\linewidth}
\caption{Bot Elo scores}
\label{tab:baseline-bots}
\vskip 0.1in
\begin{center}
\begin{small}
\begin{tabular}{llr@{}l}
    Bot & Recognized as & \multicolumn{2}{c}{\bf Elo score} \\
    \hline
    \texttt{PenumbraMixture}  & unrecognized        & $1747\, $&$\pm\, 11$ \\
    \texttt{PenumbraSimple}   & unrecognized        & $1739\, $&$\pm\, 10$ \\
    \texttt{PenumbraCache}    & unrecognized        & $1727\, $&$\pm\, 10$ \\
    \texttt{PenumbraTree}     & unrecognized        & $1641\, $&$\pm\, 9$ \\
    \texttt{StockyInference}  & \texttt{Trout}      & $1610\, $&$\pm\, 7$  \\
    \texttt{StockyInference}  & \texttt{StrangeFi.} & $1528\, $&$\pm\, 8$ \\
    \texttt{StockyInference}  & \texttt{Genetic}    & $1512\, $&$\pm\, 8$ \\
    \texttt{StockyInference}  & \texttt{StockyInf.} & $1500\, $&$\pm\, 8$ \\
    \texttt{StockyInference}  & unrecognized        & $1474\, $&$\pm\, 8$ \\
    \texttt{PenumbraNetwork}  & unrecognized        & $1376\, $&$\pm\, 9$ \\
    \texttt{AggressiveTree}   & unrecognized        & $1134\, $&$\pm\, 15$ \\
    \texttt{FullMonte}        & unrecognized        & $1028\, $&$\pm\, 20$ \\
    \texttt{Trout}            & \texttt{Trout}      & $1005\, $&$\pm\, 22$ \\
    \texttt{Trout}            & unrecognized        & $997\, $&$\pm\, 22$ \\
\end{tabular}
\end{small}
\end{center}
\end{minipage}%
\begin{minipage}{.5\linewidth}
\caption{Caution and paranoia grid search results}
\label{tab:caution-and-paranoia}
\vskip 0.1in
\begin{center}
\begin{small}
\setlength{\tabcolsep}{2pt}
\begin{tabular}{crrrrr}
    &      & \multicolumn{4}{c}{\bf Caution} \\
    &      & \multicolumn{1}{c}{0\%} & \multicolumn{1}{c}{10\%} & \multicolumn{1}{c}{20\%} & \multicolumn{1}{c}{30\%} \\
    \hline
    \parbox[t]{6px}{\multirow{4}{*}{\rotatebox[origin=c]{90}{\bf Paranoia}}} 
    & \small{ 0\%} & $1711 \spm 19$ & $1714 \spm 19$ & $1707 \spm 19$ & $1702 \spm 19$ \\
    & \small{10\%} & $1711 \spm 19$ & $1705 \spm 19$ & $1726 \spm 19$ & $1695 \spm 18$ \\
    & \small{20\%} & $1688 \spm 18$ & $1700 \spm 19$ & $1688 \spm 18$ & $1670 \spm 18$ \\
    & \small{30\%} & $1691 \spm 18$ & $1683 \spm 18$ & $1681 \spm 18$ & $1666 \spm 18$ \\
\end{tabular}
\end{small}
\end{center}
\vskip 0.15in

\caption{Exploration strategy grid search results}
\label{tab:exploration}
\vskip 0.05in
\begin{center}
\begin{small}
\begin{tabular}{lrrr}
         & \multicolumn{3}{c}{{\bf Exploration ratio} $c$}  \\
         & \multicolumn{1}{c}{1} & \multicolumn{1}{c}{2} & \multicolumn{1}{c}{4} \\
    \hline
    UCB1 &  $1698 \pm 19$ & $1686\pm 18$ & $1696 \pm 18$ \\
    aVoP &  $1696 \pm 18$ & $1680 \pm 18$ & $1695 \pm 18$ \\
\end{tabular}
\end{small}
\end{center}
\end{minipage}
\end{table}

%% file: appendix.tex
\section{Appendix}

\begin{figure*}[ht]
\centering
\begin{minipage}{1\linewidth}
\includegraphics[width=1\linewidth]{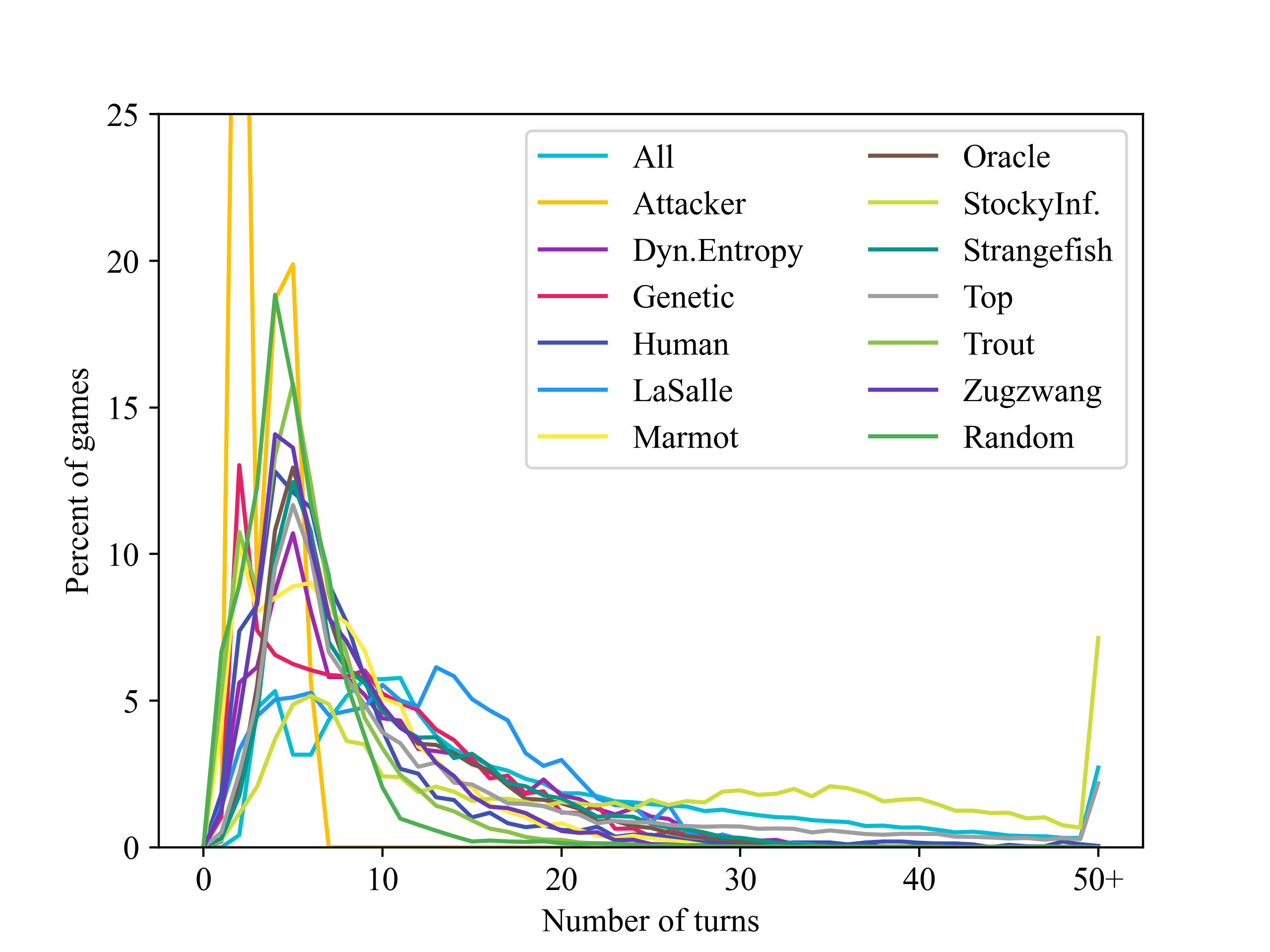}
\end{minipage}
\caption{
The historical game length distributions are shown for the
data used to train each of the headsets.
On average, the games from \focusattacker{} were the shortest,
and the games from \focusstockyinference{} where the longest.
}
\label{fig:game-length-distribution}
\end{figure*}

Table~\ref{tab:hyperparameters} lists the training hyperparameters and
runtime hyperparameters used by \texttt{PenumbraMixture}.
Table~\ref{tab:cross-eval-top-1}, Table~\ref{tab:cross-eval-top-5},
and Table~\ref{tab:cross-eval-winner-acc}
provide top-1 action, top-5 action, and winner accuracies, respectively,
between each headset in the neural network.
Figure~\ref{fig:game-length-distribution}
shows game length distributions for each headset.

The synopsis features were hand-designed.
Many of them are natural given the rules of chess.
Some of them are near duplicates of each other.
Table~\ref{tab:feature-description-1} and Table~\ref{tab:feature-description-2}
jointly provide brief descriptions of each synopsis feature plane.
These tables also include
saliency estimates averaged over five runs.
The penultimate column orders the synopsis features by
their per-bit saliency based on action gradients, and
the final column reports the average difference
of the policy head accuracies
when the model was retrained without each feature.

\begin{table}[h]
\caption{Hyperparameters used by \texttt{PenumbraMixture}}
\label{tab:hyperparameters}
\vskip 0.15in
\begin{center}
\begin{small}
\begin{tabular}{llr}
    Symbol & Parameter & Value \\
    \hline
    % \midrule
    $b$                      & Batch size                           &    $256$ \\
    $c$                      & Exploration constant                 &      $2$ \\
    $d_{\text{sense}}$       & Search depth for sense actions       &      $6$ \\
    $d_{\text{move}}$        & Search depth for move actions        &     $12$ \\
    $F$                      & \# of binary synopsis features       &    $8\stimes8\stimes104$ \\
    $k$                      & Rejection sampling persistence       &    $512$ \\
    $\ell$                   & Limited state set size               &    $128$ \\
    $m$                      & Bandit mixing constant               &      $1$ \\
    $n_{\text{particles}}$   & \# of samples to track               &   $4096$ \\
    $n_{\text{vl}}$          & Virtual loss                         &      $1$ \\
    $n_{\text{batches}}$     & Total minibatches of training        & $650000$ \\
    $n_{\text{width}}$       & Network width; \# features per layer &    $128$ \\
    $n_{\text{depth}}$       & Network depth; \# residual blocks    &    $10$ \\
    $z$                      & Depth increase threshold             & $\infty$ \\
    $\kappa$                 & Caution                              &      $0$ \\
    $\phi$                   & Paranoia                             &      $0$ \\
    $\epsilon$               & Learning rate                        & $0.0005$ \\
\end{tabular}
\end{small}
\end{center}
\vskip -0.1in
\end{table}

\subsection{2019 NeurIPS competition}

\penumbra{} was originally created to compete in the
2019 reconnaissance blind chess competition
hosted by the Conference on Neural Information Processing Systems (NeurIPS).
However, it performed very poorly in that competition,
winning fewer games than the random bot.

The program and underlying algorithm presented in this paper
are largely the same as the originals.
The main differences are that some
hyperparameters were adjusted,
the neural network was retrained with more data,
and a key bug in the \playout{} code was fixed.
Instead of choosing actions according to the policy from the neural network,
the \playout{} code erroneously always selected the last legal action.
Giving the program a \texttt{break} made a huge difference.
% literally made all the difference.

\subsection{Comparison with Kriegspiel}

A comparison between RBC and Kriegspiel chess
\citep{ciancarini2009mcts, parker2010paranoia, richards2012reasoning}
may be worthwhile. Kriegspiel chess 
also introduces uncertainty about the opposing pieces
but lacks an explicit sensing mechanism.
Instead, information is gathered solely from
captures, check notifications, and illegal move attempts.
In Kriegspiel, illegal moves are rejected and the player is allowed to choose a new
move with their increased information about the board state,
which entangles the positional and informational aspects of the game.
In contrast, sensing in RBC gives players direct control
over the amount and character of the information they possess.

Another significant difference comes from the mechanics related to check.
Capturing pieces and putting the opposing king into check
have benefits in both games:
capturing pieces leads to a material advantage,
and check often precedes checkmate.
In Kriegspiel, however, both capturing and giving check
also provide the opponent with information.
In RBC, while capturing does give the opponent information,
putting their king into check does not,
which makes sneak attacks more viable.

\subsection{Games played}

The games that were played in order to produce Table~\ref{tab:baseline-bots}, Table~\ref{tab:caution-and-paranoia}, and Table~\ref{tab:exploration}
are available for download from \url{https://github.com/w-hat/penumbra}.

\begin{table*}[ht]
\caption{Top-1 action accuracy across headsets.}
\label{tab:cross-eval-top-1}
\vspace{0.1in}
\begin{center}
\rowcolors{2}{gray!10}{white}
\setlength{\tabcolsep}{3.2pt}
\begin{tabular}{ccccccccccccccc}
    \crossevaltableheader
    \focusall{}            & \bf 33.6 & 41.5 & 40.5 & 32.3 & 43.7 & 44.1 & 41.5 & 20.3 & 31.9 & 37.1 & 36.6 & 21.9 & 41.4 & 3.8 \\
    \focustop{}            & 30.5 & \bf 43.1 & 44.5 & 32.3 & 43.9 & 44.4 & 40.3 & 19.4 & 31.6 & 22.4 & 26.0 & 18.5 & 15.6 & 3.3 \\
    \focusstrangefish{}    & 27.3 & 40.5 & \bf 45.9 & 30.3 & 36.9 & 38.8 & 33.2 & 17.9 & 27.7 & 21.0 & 23.6 & 17.5 & 14.6 & 3.3 \\
    \focuslasalle{}        & 26.0 & 34.4 & 34.6 & \bf 36.4 & 34.1 & 34.5 & 33.2 & 17.2 & 24.6 & 22.6 & 26.1 & 15.5 & 11.2 & 3.4 \\
    \texttt{Dyn.Entropy}   & 27.9 & 37.9 & 35.2 & 27.5 & \bf 50.0 & 40.1 & 37.4 & 18.0 & 32.6 & 18.3 & 22.6 & 16.9 & 13.8 & 3.3 \\
    \focusoracle{}         & 28.8 & 38.4 & 36.3 & 29.1 & 41.2 & \bf 49.3 & 35.1 & 17.2 & 29.5 & 19.9 & 26.6 & 17.0 & 11.9 & 3.4 \\
    \focuswbernar{}        & 28.8 & 38.1 & 33.9 & 30.7 & 42.9 & 38.6 & \bf 45.2 & 17.9 & 29.3 & 18.4 & 23.1 & 16.7 & 11.6 & 3.3 \\
    \focusmarmot{}         & 22.4 & 29.4 & 28.6 & 24.0 & 31.4 & 29.3 & 29.7 & \bf 24.6 & 25.0 & 16.4 & 15.5 & 15.4 & 11.1 & 3.4 \\
    \focusgenetic{}        & 24.0 & 32.5 & 30.3 & 24.1 & 39.8 & 35.0 & 32.3 & 16.9 & \bf 40.4 & 15.1 & 15.7 & 15.1 &  7.8 & 3.4 \\
    \focuszugzwang{}       & 20.5 & 21.8 & 23.2 & 20.5 & 20.4 & 23.5 & 17.3 & 11.7 & 12.3 & \bf 47.0 & 34.6 & 14.0 & 10.9 & 3.2 \\
    \focustrout{}          & 22.8 & 25.1 & 26.0 & 24.0 & 23.0 & 27.8 & 21.3 & 12.5 & 14.4 & 36.1 & \bf 41.8 & 14.9 & 14.7 & 3.7 \\
    \focushuman{}          & 23.8 & 30.1 & 30.6 & 24.9 & 31.4 & 30.1 & 28.4 & 16.8 & 24.1 & 24.7 & 24.5 & \bf 24.9 & 12.5 & 3.3 \\
    \focusattacker{}       & 10.6 & 11.7 & 11.0 &  9.8 & 11.4 & 11.6 & 12.4 &  8.6 &  8.8 &  9.2 &  9.0 &  6.7 & \bf 45.1 & 4.4 \\
    \focusrandom{}         & 14.0 & 16.7 & 16.2 & 14.0 & 16.5 & 17.8 & 16.8 &  9.4 & 11.4 & 15.7 & 16.5 & 10.4 & 10.4 & \bf 4.5 \\
\end{tabular}
\end{center}
\end{table*}

\begin{table*}[ht]
\caption{Top-5 action accuracy across headsets.}
\label{tab:cross-eval-top-5}
\vspace{0.1in}
\begin{center}
\rowcolors{2}{gray!10}{white}
\setlength{\tabcolsep}{3pt}
\begin{tabular}{ccccccccccccccc}
    \crossevaltableheader
    \focusall{}          & \bf 62.0 & 72.3 & 71.0 & 66.6 & 74.9 & 75.8 & 72.5 & 51.3 & 65.3 & 65.5 & 61.8 & 48.1 & 59.1 & 18.2 \\
    \focustop{}          & 58.8 & \bf 73.4 & 73.4 & 66.4 & 75.5 & 76.4 & 72.0 & 50.1 & 64.7 & 48.5 & 52.2 & 45.4 & 35.3 & 16.3 \\
    \focusstrangefish{}  & 56.8 & 72.2 & \bf 74.3 & 64.7 & 72.6 & 74.2 & 67.5 & 48.5 & 62.1 & 46.5 & 50.1 & 43.8 & 41.4 & 15.7 \\
    \focuslasalle{}      & 56.8 & 68.8 & 68.4 & \bf 68.2 & 69.8 & 70.6 & 68.3 & 48.7 & 58.9 & 51.5 & 56.6 & 43.6 & 30.8 & 16.8 \\
    \texttt{Dyn.Entropy} & 55.5 & 69.1 & 66.9 & 60.9 & \bf 76.7 & 72.3 & 69.6 & 47.2 & 64.4 & 38.4 & 47.2 & 42.7 & 41.4 & 16.6 \\
    \focusoracle{}       & 56.7 & 70.4 & 68.9 & 61.9 & 74.1 & \bf 77.4 & 69.1 & 45.6 & 63.4 & 43.9 & 50.6 & 43.1 & 34.4 & 16.4 \\
    \focuswbernar{}      & 57.0 & 70.0 & 67.3 & 64.7 & 74.0 & 71.3 & \bf 73.6 & 49.3 & 63.3 & 43.3 & 50.2 & 44.0 & 29.8 & 17.0 \\
    \focusmarmot{}       & 53.4 & 65.0 & 64.0 & 58.7 & 68.6 & 66.1 & 65.2 & \bf 55.4 & 59.3 & 43.0 & 46.2 & 42.6 & 32.2 & 16.2 \\
    \focusgenetic{}      & 52.6 & 65.8 & 64.2 & 58.2 & 71.3 & 69.7 & 65.8 & 45.4 & \bf 70.5 & 37.4 & 41.6 & 40.9 & 27.0 & 16.2 \\
    \focuszugzwang{}     & 42.8 & 45.0 & 45.4 & 46.0 & 42.1 & 47.5 & 42.4 & 32.5 & 32.8 & \bf 71.6 & 57.9 & 35.3 & 29.0 & 15.4 \\
    \focustrout{}        & 49.3 & 54.5 & 54.4 & 53.9 & 53.7 & 57.2 & 52.5 & 38.4 & 42.2 & 62.9 & \bf 63.4 & 38.7 & 40.9 & 18.1 \\
    \focushuman{}        & 53.5 & 62.9 & 62.4 & 58.3 & 64.7 & 64.6 & 62.5 & 45.9 & 55.7 & 53.7 & 53.4 & \bf 51.9 & 33.4 & 16.3 \\
    \focusattacker{}     & 35.2 & 39.5 & 38.8 & 34.7 & 40.8 & 40.2 & 39.7 & 31.1 & 33.8 & 33.0 & 32.4 & 28.4 & \bf 61.9 & 20.0 \\
    \focusrandom{}       & 39.7 & 45.6 & 44.5 & 41.2 & 46.4 & 47.9 & 46.0 & 31.8 & 37.7 & 41.3 & 42.1 & 31.5 & 30.3 & \bf 20.7 \\
\end{tabular}
\end{center}
\end{table*}

\begin{table*}[ht]
\caption{Winner accuracy across headsets.}
\label{tab:cross-eval-winner-acc}
\vspace{0.1in}
\begin{center}
\rowcolors{2}{gray!10}{white}
\setlength{\tabcolsep}{3pt}
\begin{tabular}{ccccccccccccccc}
    \crossevaltableheader
    \focusall{}          & \bf 74.8 & 73.4 & 76.6 & 68.4 & 74.6 & 76.3 & 67.8 & 79.3 & 74.1 & 76.7 & 76.7 & 72.1 & 79.6 & 91.3 \\
    \focustop{}          & 63.1 & \bf 82.4 & \bf 86.7 & 68.0 & 76.0 & 80.6 & 66.3 & 65.6 & 73.2 & 49.1 & 55.1 & 52.7 & 47.7 & 29.9 \\
    \focusstrangefish{}  & 64.1 & 82.1 & 86.6 & 67.5 & 76.2 & 80.6 & 65.2 & 65.8 & 72.7 & 50.1 & 55.4 & 55.8 & 60.9 & 37.3 \\
    \focuslasalle{}      & 69.9 & 77.3 & 80.7 & \bf 69.7 & 76.0 & 78.8 & 67.1 & 73.5 & 75.9 & 65.1 & 68.3 & 62.4 & 71.9 & 60.7 \\
    \texttt{Dyn.Entropy} & 67.8 & 80.4 & 85.0 & 69.4 & \bf 78.9 & 80.8 & 67.0 & 71.7 & 75.6 & 58.5 & 61.1 & 61.5 & 65.0 & 47.3 \\
    \focusoracle{}       & 66.0 & 82.0 & 86.4 & 69.4 & 77.3 & \bf 81.4 & 66.9 & 69.2 & 75.3 & 53.5 & 58.2 & 57.9 & 59.5 & 39.7 \\
    \focuswbernar{}      & 71.3 & 78.5 & 82.5 & 69.3 & 77.2 & 79.5 & \bf 68.3 & 75.3 & 76.8 & 64.0 & 65.8 & 66.9 & 72.6 & 68.9 \\
    \focusmarmot{}       & 70.6 & 67.7 & 70.1 & 64.5 & 70.9 & 72.3 & 65.7 & \bf 80.8 & 72.9 & 73.9 & 73.4 & 69.2 & 77.0 & 72.6 \\
    \focusgenetic{}      & 67.1 & 80.0 & 84.1 & 68.6 & 77.1 & 80.3 & 67.1 & 71.5 & \bf 77.9 & 56.5 & 60.8 & 60.9 & 62.1 & 44.3 \\
    \focuszugzwang{}     & 68.3 & 54.6 & 54.0 & 59.7 & 61.4 & 60.0 & 61.7 & 76.1 & 64.4 & \bf 80.1 & 78.8 & 72.5 & 78.5 & 88.9 \\
    \focustrout{}        & 69.7 & 58.6 & 59.1 & 60.9 & 63.6 & 64.4 & 63.4 & 77.9 & 69.0 & 77.9 & \bf 79.8 & 72.4 & 78.6 & 87.3 \\
    \focushuman{}        & 71.1 & 64.8 & 66.0 & 63.8 & 67.7 & 68.0 & 65.1 & 76.6 & 70.4 & 74.6 & 76.2 & \bf 73.3 & 77.7 & 90.1 \\
    \focusattacker{}     & 63.7 & 47.1 & 46.1 & 55.6 & 53.8 & 50.9 & 56.5 & 71.4 & 59.5 & 75.0 & 73.4 & 71.5 & \bf 80.1 & 92.7 \\
    \focusrandom{}       & 51.0 & 25.5 & 20.9 & 43.4 & 34.8 & 28.5 & 46.5 & 54.9 & 38.2 & 65.7 & 56.6 & 62.5 & 69.5 & \bf 94.7 \\
\end{tabular}
\end{center}
\end{table*}

\begin{table*}[ht]
\caption{%
Synopsis feature descriptions, saliency estimates, and ablation study results.
}
\label{tab:feature-description-1}
\vspace{0.1in}
\begin{center}
\begin{tabular}{clrrrrrr}
    \featuretableheader
    0 &  East side (constant)      & 3.24 & 0.86 & 0.80 & 0.91 & 34 & 0.27 \\
    1 &  West side (constant)      & 3.12 & 0.82 & 0.84 & 0.81 & 43 & 0.00 \\
    2 &  South side (constant)     & 3.24 & 0.86 & 0.78 & 0.94 & 33 & -0.08 \\
    3 &  North side (constant)     & 3.22 & 0.82 & 0.89 & 0.75 & 44 & 0.27 \\
    4 &  Rank 1 (constant)         & 3.15 & 0.80 & 0.81 & 0.76 & 50 & -0.03 \\
    5 &  Rank 8 (constant)         & 3.12 & 0.82 & 0.85 & 0.61 & 42 & -0.07 \\
    6 &  A-file (constant)         & 3.04 & 0.78 & 0.81 & 0.53 & 59 & 0.18 \\
    7 &  H-file (constant)         & 3.08 & 0.80 & 0.83 & 0.58 & 51 & 0.15 \\
    8 &  Dark squares (constant)   & 3.08 & 0.82 & 0.81 & 0.82 & 45 & 0.21 \\
    9 &  Light squares (constant)  & 3.03 & 0.78 & 0.77 & 0.78 & 61 & 0.01 \\
    10 &  Stage (move or sense)    & 7.80 & 3.14 & 2.82 & 3.45 & 0 & -0.19 \\
    11 &  Not own piece            & 5.40 & 1.43 & 2.66 & 1.13 & 8 & -0.29 \\
    12 &  Own pawns                & 4.16 & 1.14 & 1.07 & 1.73 & 14 & 0.01 \\
    13 &  Own knights              & 3.68 & 0.93 & 0.91 & 1.63 & 22 & 0.09 \\
    14 &  Own bishops              & 3.46 & 0.89 & 0.87 & 1.63 & 27 & 0.02 \\
    15 &  Own rooks                & 3.67 & 0.94 & 0.93 & 1.12 & 21 & 0.03 \\
    16 &  Own queens               & 3.28 & 0.87 & 0.85 & 2.24 & 32 & 0.06 \\
    17 &  Own king                 & 3.14 & 0.79 & 0.79 & 0.88 & 52 & 0.10 \\
    18 &  Definitely not opposing pieces  & 3.85 & 1.14 & 1.03 & 1.21 & 13 & -0.10 \\
    19 &  Definitely opposing pawns       & 3.49 & 1.01 & 1.02 & 0.73 & 17 & -0.08 \\
    20 &  Definitely opposing knights     & 3.30 & 0.93 & 0.93 & 0.61 & 23 & -0.05 \\
    21 &  Definitely opposing bishops     & 3.21 & 0.88 & 0.88 & 0.59 & 29 & -0.02 \\
    22 &  Definitely opposing rooks       & 3.04 & 0.81 & 0.82 & 0.38 & 47 & 0.02 \\
    23 &  Definitely opposing queens      & 3.15 & 0.85 & 0.85 & 0.60 & 35 & -0.10 \\
    24 &  Definitely opposing king        & 3.60 & 0.92 & 0.91 & 2.27 & 26 & 0.04 \\
    25 &  Possibly not opposing pieces    & 5.22 & 1.54 & 1.34 & 1.56 & 5 & -0.04 \\
    26 &  Possibly opposing pawns         & 3.50 & 0.92 & 0.92 & 0.93 & 24 & 0.06 \\
    27 &  Possibly opposing knights       & 2.97 & 0.77 & 0.77 & 0.81 & 67 & 0.07 \\
    28 &  Possibly opposing bishops       & 2.95 & 0.75 & 0.74 & 0.89 & 70 & 0.09 \\
    29 &  Possibly opposing rooks         & 3.01 & 0.75 & 0.76 & 0.63 & 69 & -0.18 \\
    30 &  Possibly opposing queens        & 3.05 & 0.78 & 0.77 & 1.05 & 57 & -0.07 \\
    31 &  Possibly opposing kings         & 4.86 & 1.48 & 1.43 & 2.64 & 7 & -0.04 \\
    32 &  Last from                       & 2.77 & 0.72 & 0.72 & 0.83 & 76 & -0.11 \\
    33 &  Last to                         & 3.28 & 0.96 & 0.96 & 1.40 & 19 & 0.02 \\
    34 &  Last own capture                & 3.10 & 0.83 & 0.83 & 1.17 & 40 & 0.07 \\
    35 &  Last opposing capture           & 8.04 & 2.83 & 2.82 & 6.51 & 1 & -0.08 \\
    36 &  Definitely attackable           & 2.72 & 0.70 & 0.62 & 0.78 & 84 & -0.06 \\
    37 &  Definitely attackable somehow   & 2.73 & 0.71 & 0.65 & 0.78 & 80 & -0.02 \\
    38 &  Possibly attackable             & 3.02 & 0.81 & 0.71 & 0.92 & 48 & 0.19 \\
    39 &  Definitely doubly attackable    & 2.67 & 0.66 & 0.63 & 0.80 & 92 & -0.11 \\
    40 &  Definitely doubly attackable somehow  & 2.66 & 0.69 & 0.67 & 0.80 & 88 & 0.14 \\
    41 &  Possibly doubly attackable            & 2.71 & 0.75 & 0.73 & 0.83 & 71 & -0.26 \\
    42 &  Definitely attackable by pawns        & 3.54 & 0.92 & 0.92 & 2.38 & 25 & 0.13 \\
    43 &  Possibly attackable by pawns          & 3.11 & 0.78 & 0.78 & 0.95 & 58 & -0.10 \\
    44 &  Definitely attackable by knights      & 2.91 & 0.72 & 0.71 & 0.84 & 77 & 0.24 \\
    45 &  Definitely attackable by bishops      & 2.60 & 0.64 & 0.61 & 0.80 & 95 & 0.15 \\
    46 &  Possibly attackable by bishops        & 2.60 & 0.68 & 0.64 & 0.85 & 89 & -0.07 \\
    47 &  Definitely attackable by rooks        & 2.63 & 0.65 & 0.64 & 0.75 & 93 & 0.07 \\
    48 &  Possibly attackable by rooks          & 2.74 & 0.70 & 0.69 & 0.77 & 81 & 0.00 \\
    49 &  Possibly attackable without king      & 2.72 & 0.70 & 0.63 & 0.79 & 82 & 0.19 \\
    50 &  Possibly attackable without pawns     & 2.63 & 0.67 & 0.62 & 0.73 & 90 & 0.17 \\
    51 &  Definitely attackable by opponent     & 3.25 & 0.87 & 0.91 & 0.77 & 31 & -0.03 \\
\end{tabular}
\end{center}
\end{table*}

\begin{table*}[ht]
\caption{%
Synopsis feature descriptions, saliency estimates, and ablation study results (continued).
}
\label{tab:feature-description-2}
\vspace{0.1in}
\begin{center}
\begin{tabular}{clrrrrrr}
    \featuretableheader
    52 &  Possibly attackable by opponent        & 3.15 & 0.84 & 0.90 & 0.81 & 37 & 0.06 \\
    53 &  Definitely doubly attackable by opp.   & 2.56 & 0.65 & 0.66 & 0.56 & 94 & -0.01 \\
    54 &  Possibly doubly attackable by opp.     & 2.67 & 0.71 & 0.73 & 0.66 & 79 & -0.13 \\
    55 &  Definitely attackable by opp. pawns    & 3.10 & 0.87 & 0.87 & 1.69 & 30 & 0.12 \\
    56 &  Possibly attackable by opp. pawns      & 2.84 & 0.77 & 0.77 & 1.10 & 62 & 0.30 \\
    57 &  Definitely attackable by opp. knights  & 2.78 & 0.69 & 0.70 & 0.59 & 87 & 0.21 \\
    58 &  Possibly attackable by opp. knights    & 2.14 & 0.52 & 0.51 & 0.55 & 102 & 0.08 \\
    59 &  Definitely attackable by opp. bishops  & 2.66 & 0.67 & 0.67 & 0.63 & 91 & 0.09 \\
    60 &  Possibly attackable by opp. bishops    & 2.16 & 0.53 & 0.52 & 0.56 & 101 & -0.09 \\
    61 &  Definitely attackable by opp. rooks    & 2.77 & 0.70 & 0.72 & 0.48 & 83 & -0.19 \\
    62 &  Possibly attackable by opp. rooks      & 2.10 & 0.51 & 0.53 & 0.47 & 103 & 0.20 \\
    63 &  Possibly attackable by opp. w/o king   & 2.55 & 0.64 & 0.63 & 0.64 & 96 & -0.17 \\
    64 &  Possibly attackable by opp. w/o pawns  & 2.45 & 0.62 & 0.61 & 0.62 & 97 & -0.13 \\
    65 &  Possibly safe opposing king            & 6.04 & 2.06 & 2.01 & 3.38 & 2 & 0.07 \\
    66 &  Squares the opponent may move to       & 2.40 & 0.60 & 0.60 & 0.60 & 98 & 0.01 \\
    67 &  Possible castle state for opponent     & 3.09 & 0.79 & 0.79 & 0.72 & 53 & 0.00 \\
    68 &  Known squares                          & 4.94 & 1.52 & 1.67 & 1.45 & 6 & 0.13 \\
    69 &  Own king's king-neighbors              & 3.10 & 0.78 & 0.77 & 0.93 & 56 & 0.14 \\
    70 &  Own king's knight-neighbors            & 2.82 & 0.71 & 0.70 & 0.91 & 78 & 0.31 \\
    71 &  Definitely opp. knights near king      & 3.09 & 0.79 & 0.79 & 1.64 & 54 & 0.13 \\
    72 &  Possibly opp. knights near king        & 5.13 & 1.72 & 1.72 & 2.77 & 4 & -0.01 \\
    73 &  Own king's bishop-neighbors            & 2.74 & 0.69 & 0.68 & 0.86 & 85 & -0.10 \\
    74 &  Definitely opp. bishops near king      & 3.04 & 0.79 & 0.79 & 0.89 & 55 & 0.23 \\
    75 &  Possibly opp. bishops near king        & 5.23 & 1.75 & 1.75 & 2.41 & 3 & -0.11 \\
    76 &  Own king's rook-neighbors              & 2.76 & 0.69 & 0.68 & 0.83 & 86 & -0.13 \\
    77 &  Definitely opp. rooks near king        & 3.10 & 0.81 & 0.81 & 0.87 & 49 & 0.31 \\
    78 &  Possibly opp. rooks near king          & 4.45 & 1.40 & 1.40 & 1.55 & 10 & 0.05 \\
    79 &  All own pieces                         & 5.26 & 1.36 & 1.09 & 2.47 & 11 & -0.01 \\
    80 &  Definitely empty squares               & 3.69 & 0.96 & 1.05 & 0.84 & 20 & -0.13 \\
    81 &  May castle eventually                  & 3.11 & 0.81 & 0.81 & 1.26 & 46 & 0.24 \\
    82 &  Possibly may castle                    & 3.05 & 0.77 & 0.77 & 0.63 & 68 & 0.05 \\
    83 &  Definitely may castle                  & 3.04 & 0.77 & 0.77 & 0.87 & 66 & 0.12 \\
    84 &  Own queens' rook-neighbors             & 2.20 & 0.54 & 0.53 & 0.63 & 100 & 0.04 \\
    85 &  Own queens' bishop-neighbors           & 2.33 & 0.57 & 0.57 & 0.67 & 99 & 0.06 \\
    86 &  Previous definitely not opp. pieces    & 3.82 & 0.88 & 0.87 & 0.89 & 28 & -0.30 \\
    87 &  Previous definitely opp. pawns         & 4.16 & 1.16 & 1.18 & 0.88 & 12 & 0.14 \\
    88 &  Previous definitely opp. knights       & 3.02 & 0.77 & 0.77 & 0.72 & 63 & 0.12 \\
    89 &  Previous definitely opp. bishops       & 2.92 & 0.73 & 0.73 & 0.70 & 75 & -0.02 \\
    90 &  Previous definitely opp. rooks         & 3.60 & 1.01 & 1.02 & 0.56 & 16 & 0.05 \\
    91 &  Previous definitely opp. queens        & 3.93 & 1.11 & 1.11 & 1.00 & 15 & 0.22 \\
    92 &  Previous definitely opp. king          & 3.33 & 0.83 & 0.82 & 1.58 & 38 & -0.04 \\
    93 &  Previous possibly not opp. pieces      & 4.40 & 1.43 & 1.21 & 1.47 & 9 & -0.05 \\
    94 &  Previous possibly opp. pawns           & 3.27 & 0.83 & 0.82 & 0.92 & 39 & 0.21 \\
    95 &  Previous possibly opp. knights         & 3.04 & 0.78 & 0.78 & 0.73 & 60 & 0.22 \\
    96 &  Previous possibly opp. bishops         & 3.10 & 0.74 & 0.74 & 0.78 & 73 & 0.16 \\
    97 &  Previous possibly opp. rooks           & 2.94 & 0.77 & 0.79 & 0.45 & 64 & 0.10 \\
    98 &  Previous possibly opp. queens          & 3.02 & 0.74 & 0.74 & 0.81 & 74 & -0.07 \\
    99 &  Previous possibly opp. king            & 3.14 & 0.83 & 0.82 & 1.15 & 41 & 0.04 \\
    100 &  Previous last from                    & 2.85 & 0.75 & 0.74 & 0.85 & 72 & -0.04 \\
    101 &  Previous last to                      & 3.36 & 1.00 & 1.00 & 1.50 & 18 & 0.21 \\
    102 &  Previous own capture                  & 3.05 & 0.84 & 0.84 & 1.13 & 36 & -0.09 \\
    103 &  Previous opposing capture             & 2.93 & 0.77 & 0.77 & 1.05 & 65 & 0.05 \\
\end{tabular}
\end{center}
\end{table*}